%% file: arxiv.tex
\documentclass{article}

\PassOptionsToPackage{numbers,compress}{natbib}

\usepackage[preprint]{neurips_2025}

\usepackage[utf8]{inputenc} 
\usepackage[T1]{fontenc}    
\usepackage{hyperref}       
\usepackage{url}            
\usepackage{booktabs}       
\usepackage{amsfonts}       
\usepackage{nicefrac}       
\usepackage{microtype}      
\usepackage{xcolor}         
\usepackage{xspace}
\usepackage{graphicx}
\usepackage{tikz}
\usepackage{pgf-pie}
\usepackage{pgfplots}
\pgfplotsset{compat=1.18}
\usepackage{pifont}
\usepackage{capt-of}
\usepackage{enumitem}
\usepackage{amsmath,amssymb,amsbsy,dsfont,bm,bbm,mathrsfs,mathtools}
\usepackage{algorithm,algpseudocode,listings}
\usepackage{multirow,adjustbox,diagbox,threeparttable,tabularray}
\usepackage{colortbl}
\usepackage{wrapfig,lipsum}
\usepackage[capitalize]{cleveref}  
\crefname{section}{Sec.}{Secs.}
\Crefname{section}{Section}{Sections}
\crefname{table}{Tab.}{Tabs.}
\Crefname{table}{Table}{Tables}
\crefname{figure}{Fig.}{Figs.}
\Crefname{figure}{Figure}{Figures}
\crefname{equation}{Eq.}{Eqs.}
\Crefname{equation}{Equation}{Equations}
\newcommand{\method}{\texttt{OmniTry}\xspace}

\newcommand{\tablestyle}[2]{\setlength{\tabcolsep}{#1}\renewcommand{\arraystretch}{#2}\centering\footnotesize}

\title{OmniTry: Virtual Try-On Anything without Masks}

\makeatletter
\def\@fnsymbol#1{%
  \ifcase#1\or \dagger \or \daggerdbl \or \asteriskcentered \or \section \or \paragraph \or \bardbl \fi
}
\makeatother

\author{%
  Yutong Feng$^{1,}$\thanks{Project Leader. $^{\ddagger}$Corresponding Author.} \\
  \And
  Linlin Zhang$^2$ \\
  \And
  Hengyuan Cao$^2$ \\
   \And
  Yiming Chen$^1$ \\
  \AND
  \vspace{-6pt}
  Xiaoduan Feng$^1$ \\
  \And
  Jian Cao$^1$ \\
  \And 
  Yuxiong Wu$^1$ \\
  \And
  Bin Wang$^{1,\ddagger}$ \\
  \AND
  $^1 \textnormal{Kunbyte AI}$
  \hspace{0.3em} 
  $^2 \textnormal{Zhejiang University}$
  \AND
  \texttt{\{fengyutong.fyt, binwang393\}@gmail.com}
  \hspace{0.3em} 
  \texttt{\{zhanglinlinlin, caohy\}@zju.edu.cn} \\
  \texttt{\{chenyiming, fengxiaoduan, caojian, wuyuxiong\}@k-fashionshop.com}
}

\begin{document}

\maketitle

\input{sections/0_abstract}

\input{sections/1_introduction}
\input{sections/2_related}

\input{sections/3_method}
\input{sections/4_experiment}
\input{sections/5_conclusion}

\begingroup
\small
\bibliographystyle{plainnat}
\bibliography{references}
\endgroup








\newpage
\appendix

\input{sections_appendix/benchmark_details}

\input{sections_appendix/train_dataset_details}

\input{sections_appendix/model_details}
\input{sections_appendix/comp_methods}
\input{sections_appendix/more_vis}





\end{document}

%% file: sections/0_abstract.tex
\begin{figure}[ht]
    \centering
    \includegraphics[width=0.92 \textwidth]{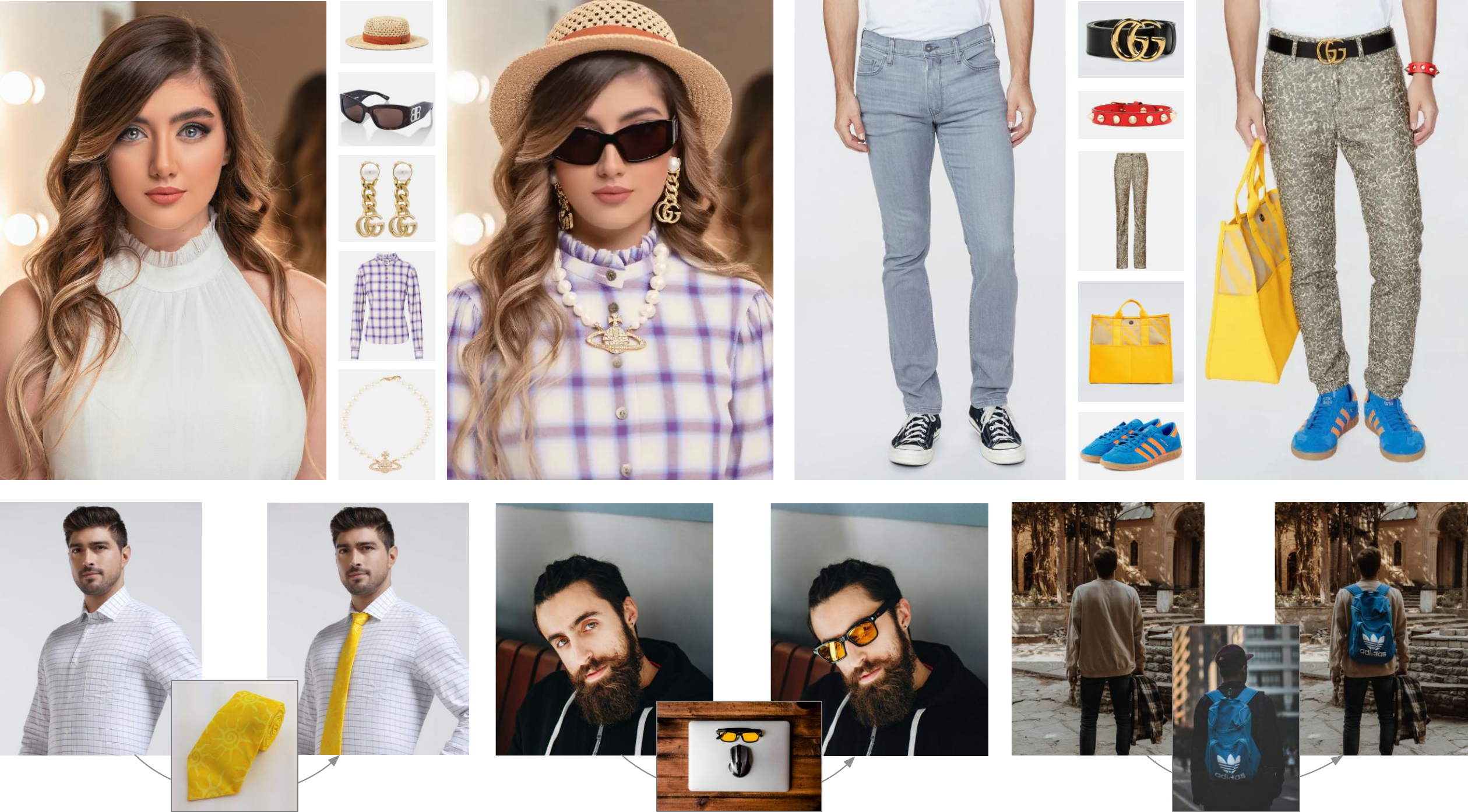}
    \vspace{-4pt}
     \caption{Try-on results of various wearable objects generated by \method, which supports object images with white or natural backgrounds, and even try-on results as input.}
    \label{fig:teaser}
\end{figure}

\begin{abstract}

Virtual Try-ON (VTON) is a practical and widely-applied task, for which most of existing works focus on clothes.
This paper presents \method, a unified framework that extends VTON beyond garment to encompass any wearable objects, \textit{e.g.}, jewelries and accessories, with mask-free setting for more practical applications.
%
%
When extending to various types of objects, data curation is challenging for obtaining paired images, \textit{i.e.,} the object image and the corresponding try-on result.
To tackle this problem, we propose a two-staged pipeline:
%
For the first stage, we leverage large-scale unpaired images, \textit{i.e.,} portraits with any wearable items, to train the model for mask-free localization.
Specifically, we repurpose the inpainting model to automatically draw objects in suitable positions given an empty mask.
%
For the second stage, the model is further fine-tuned with paired images to transfer the consistency of object appearance.
We observed that the model after the first stage shows quick convergence even with few paired samples.
\method is evaluated on a comprehensive benchmark consisting of 12 common classes of wearable objects, with both in-shop and in-the-wild images. 
Experimental results suggest that \method shows better performance on both object localization and ID-preservation compared with existing methods.
%
The code, model weights, and evaluation benchmark of \method are available at \url{https://omnitry.github.io/}.
\end{abstract}

%% file: sections/1_introduction.tex
\section{Introduction}

The image-based virtual try-on (VTON)~\cite{han2018viton} has received tremendous attention due to its wide application in e-commerce.
Given a person image and a garment image, the purpose of VTON is to transfer the garment onto the person as a preview.
Thanks to the success of large-scale image generative models~\cite{stablediffusion,sdxl,sd3,flux2024} with their photorealistic aesthetics, recent efforts~\cite{xu2025ootdiffusion,kim2024stableviton,idmvton,chong2024catvton,zhang2024boow} have achieved satisfying performance on both generation quality and garment identity preservation.

Despite the advancement of VTON, existing methods mainly concentrate on clothing try-on.
Though some works have explored the extension to non-clothing, such as shoes~
\cite{chou2019pivtons} and ornaments~\cite{miao2025shining}, 
there still lacks a unified framework in the literature, supporting any types of wearable objects.
Furthermore, most methods require the indication of wearing area on person (\textit{e.g.,} masks or bounding boxes), or use automatic human-body parsers~\cite{human_parsing_survey} to identify the area.
When extending to anything try-on, it would be impractical to expect users to draw the targeting area, as the interaction between the model and various objects can be more considerably more complex.
It is also challenging to leverage existing parsers to detect appropriate try-on areas for diverse objects.
Thus, we follow the mask-free setting~\cite{issenhuth2020not,ge2021parser,zhang2024boow} for the model to automatically localize the area with natural composition. 

When confronting anything try-on, one key challenge is the data collection. 
Generally, the training of VTON requires large-scale \textit{paired} images, consisting of a single-shot of the garment, and a corresponding person try-on result.
Most datasets are curated from e-commerce websites, with at least thousands of samples, \textit{e.g.,} VITON-HD~\cite{choi2021vitonhd} and DressCode~\cite{morelli2022dresscode}.
While for many common types of wearable objects, such as hats and ties, there is no abundant quantity of paired data, but only the product pictures.
This limitation makes it necessary to develop an efficient training framework.

In this paper, we present \method, targeting mask-free virtual try-on for any wearable object.
\method reduces the heavy reliance on paired training samples, leveraging large-scale \textit{unpaired} images for prior learning.
The unpaired images refer to the image containing a person with any wearable objects, which can be easily obtained from existing database.
The training of \method can be separated into two stages:
(i) The first stage is completely conducted on unpaired data.
We use multi-modal large language models (MLLMs)~\cite{bai2025qwen2} to list all wearable items with descriptions. 
Each item is detected and erased from the image, forming a training pair. 
Then an image generative model is trained to re-paint the item, prompted by the corresponding text description.
After stage one, the model is expected to know how to transfer various objects onto the person in proper position, size and orientation.
(ii) For the second stage, we further leverage high-quality paired data to fine-tune the model.
Object image is introduced into the context, modulating the model to preserve the consistency of object appearance. 
Building upon the model from stage one, we observe that ID-consistency is quickly adapted even fine-tuned with few samples.
To summarize, the two stages in \method contributes the ability of mask-free localization and ID-preservation, respectively.

Regarding the model design, we leverage the diffusion transformer as backbone, and compare two variants, \textit{i.e.,} text-to-image and inpainting model.
Experimental results show that the inpainting model can be rapidly repurposed as a mask-free generative model, by simply setting the mask input with all-zero values.
Image tokens from different images are concatenated in the sequence dimension, and processed with full-attention mechanism for consistency learning~\cite{omnicontrol,easycontrol,DBLP:24UniReal,iclora}.
We employ efficient adapter tuning techniques for transferring the model to this task.
More specifically, we implement two distinct adapters that handle the tokens from person and object images, individually.

The erasure of wearable object is observed with critical impact.
A naive solution is to call object-removal models~\cite{lama,powerpaint,smarteraser} to fill the area of objects.
However, we notice that while the processed area appears visually normal, it contains imperceptible artifacts.
Thus, the model learn undesirable shortcuts by identifying these traces, resulting in poor generalization to natural images.
To tackle this problem, we propose \textit{traceless erasing} to eliminate the artifacts.
We conduct image-to-image~\cite{meng2021sdedit} to subtly re-paint the entire image after erasure.
Subsequently, the original try-on image is blended with the re-painted image, ensuring the non-object area remains unchanged.
Traceless erasing disrupts the erasure boundaries, thereby compelling the model to learn genuine try-on capability.

We construct a comprehensive evaluation benchmark covering 12 common types of wearable objects, divided into clothes, shoes, jewelries and accessories.
To fully investigate the model robustness, the objects are set on white and natural backgrounds, or try-on images, referring to ~\cref{fig:teaser}.
Metrics are designed to evaluate the object consistency, person preservation and wearing position.
Experimental results indicate that \method outperforms existing methods, and achieves efficient few-shot training.

%% file: sections/2_related.tex
\section{Related Works}

\paragraph{Controllable Image Generation.}
The breakthrough of diffusion model~\cite{ddpm} has driven extensive research on controllable image generation. 
ControlNet~\cite{DBLP:23ControlNet} and related pioneering works~\cite{DBLP:24t2iAdapter,DBLP:23UniControl,DBLP:23UniControlNet} explore precise control with diverse conditions. 
IP-Adapter~\cite{2023IP} and related studies~\cite{DBLP:23OneWord,DBLP:2023DreamTuner,DBLP:23MultiConcept,DBLP:23BLIPdiffusion,DBLP:24SubjectDiffusion} investigate online concept control to achieve subject customized generation.
Recent developments in DiT~\cite{dit} have further propelled generalized image generation and editing. 
In-context LoRA~\cite{iclora} enables diverse thematic generation with image concatenation.
OminiControl~\cite{omnicontrol} introduces task-agnostic condition control with minimal model modification.
OmniGen~\cite{DBLP:24OmniGen} unifies multi-task processing via large vision-language models.
UniReal~\cite{DBLP:24UniReal} achieves unified image editing via full-attention and video data prior.
VisualCloze~\cite{li2025visualclozeuniversalimagegeneration} enhances visual in-context learning for cross-domain generalization.
For localized image customization, Anydoor~\cite{chen2024anydoor} pioneers to transfer subject into specified region.
MimicBrush~\cite{DBLP:24Mimicbrush} extends to local components transferring with imitative editing. 
ACE++~\cite{2025ACE} establishes a unified paradigm for generation and editing tasks. 

\paragraph{Image-based Virtual Try-On (VTON)} has emerged as a critical task attracting tremendous efforts.
VITON ~\cite{han2018viton} introduces Thin Plate Spline transformations~\cite{DBLP:journals/pami/Bookstein89} for multi-stage garment processing.
CP-VTON~\cite{2018Toward} formalizes explicit geometric warping and texture synthesis stages. 
GP-VTON~\cite{DBLP:23GPvton} combines local flow estimation with global parsing to improve detail preservation. 
These warping-based approaches, however, face persistent challenges in cross-sample alignment and generalization. 
This motivates the adoption of diffusion models~\cite{ddpm}, including TryOnDiffusion's parallel U-Net~\cite{DBLP:23TryOnDiffusion}, LADIVTON's garment tokenization ~\cite{DBLP:23LaDIvton}, and DCI-VTON's hybrid warping-diffusion framework~\cite{DBLP:23Taming}. 
OOTDiffusion~\cite{xu2025ootdiffusion} and FitDiT~\cite{DBLP:24FitDiT} enhance detail fidelity through specialized attention mechanisms.
Though with advanced results, most of them remain constrained by intensive preprocessing requirements (\textit{e.g.,} wearing masks and pose estimation). 
Boow-VTON~\cite{zhang2024boow} creates a mask-free approach through in-the-wild data augmentation.
Any2AnyTryon~\cite{DBLP:25Any2AnyTryon} pioneers fully mask-free implementations, eliminating dependency on masks or poses.


%% file: sections/3_method.tex
\section{Method}

\subsection{Preliminary}
\label{chap:3_1_preliminary}
\textbf{Diffusion Transformer (DiT).}  
\method is developed on DiT~\cite{dit}, a scalable transformer architecture for diffusion-based generation.
The image is encoded into latent space through an autoencoder~\cite{vae}, and patchified into tokens~\cite{vit}. 
Diffusion process~\cite{ddpm} is conducted on tokens with a transformer consuming the noisy tokens and predicts for denoising.
Recent advancement in DiT, \textit{i.e.,} recified flow matching~\cite{liu2022flow} and rotary position embedding (RoPE)~\cite{rope}, are also involved in this paper. 

\textbf{Virtual Try-On (VTON).} 
Given a person image $\mathcal{I}_P$ and a wearable object image $\mathcal{I}_O$, the try-on result image is noted as $\mathcal{I}_T$. 
Suppose the segmentation mask of the object in $\mathcal{I}_T$ is $\mathcal{M}$, then the target of VTON is three-fold: (i) the consistency between objects in original and try-on images, \textit{i.e.} $\min \text{similarity}(\mathcal{I}_T  \mathcal{M}, ~\mathcal{I}_O)$, (ii) the preservation of non-wearing areas, \textit{i.e.,} $\mathcal{I}_T (1 -\mathcal{M}) = \mathcal{I}_P (1 -\mathcal{M})$, (iii) the object is properly located on person, evaluated through the quality of $\mathcal{I}_T$. 

\subsection{Stage-1: Mask-Free Localization}
\begin{figure}[t]
    \centering
    \includegraphics[width=0.98 \textwidth]{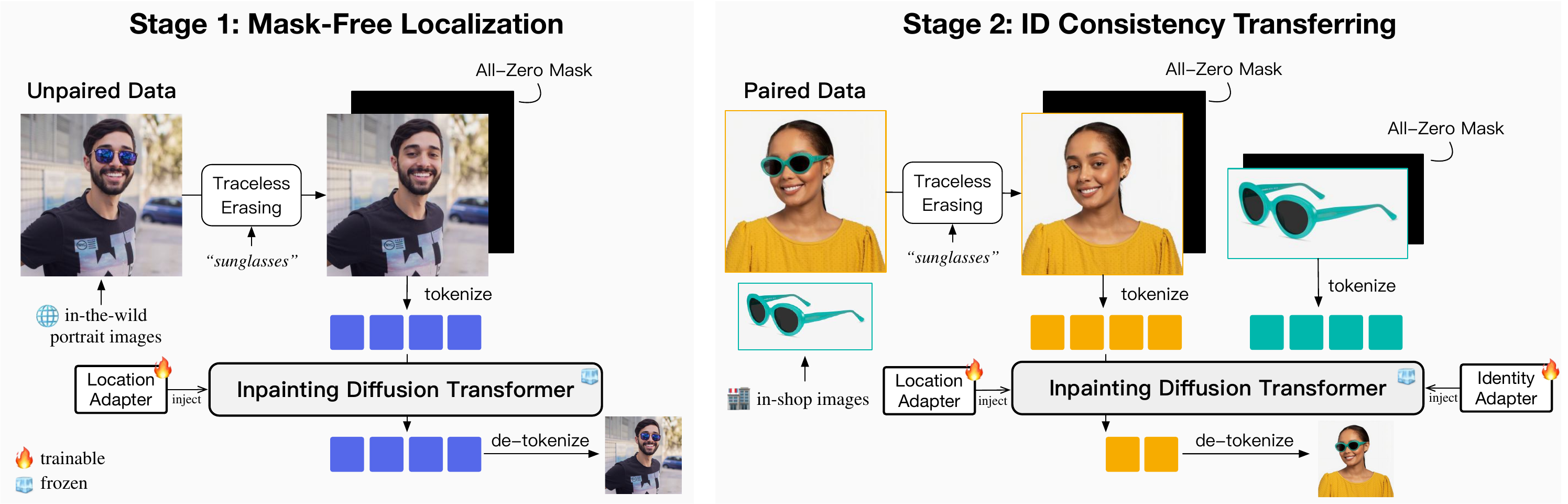}
    \vspace{-8pt}
     \caption{\textbf{The two-staged training pipeline of \method.} The first stage is built on in-the-wild portrait images to add wearable object onto the person in mask-free manner. The second stage introduces in-shop paired images, and targets to control the consistency of object appearance.}
         \vspace{-10pt}
    \label{fig:pipeline}
\end{figure}

As illustrated in\cref{fig:pipeline}, the training of \method consists of two stages, corresponding to the abilities of localization and ID-preservation, respectively. 
In the first stage, the objective of training can be regarded as ``garment-free VTON'', in contrast to the ``model-free VTON'' in the literature~\cite{DBLP:25Any2AnyTryon}. 
Given the person image $\mathcal{I}_P$ and the object description, the model aims to edit $\mathcal{I}_P$ by adding the object as described. The type and detailed appearance of object are only prompted by input text. 
Control signal indicating the wearing area, \textit{e.g.,} bounding boxes, masks or selecting point, is not introduced here.
Such an objective enforces the model to concentrate on \textit{where} to paint the object, and \textit{how} to blend it harmoniously with the person image.
The training of stage one can be easily supervised by a portrait image database, for which we introduce how to construct the training samples in the next paragraph.

\textbf{Unpaired Data Pre-process.} We refer single portrait image as \textit{unpaired} image with only try-on result $\mathcal{I}_T$, in contrast to the paired images $(\mathcal{I}_T, \mathcal{I}_O)$ in next stage. 
We start by curating a large-scale dataset containing any human-related images. The dataset is filtered by a classifier for images with a person wearing at least one object.
Following that, we leverage a MLLM, Qwen-VL 2.5~\cite{bai2025qwen2}, to list all potential wearable objects in each image. 
The output includes both the type of object and its appearance description.
We also prompt MLLM to add an interaction description, \textit{e.g.,} ``wearing sunglasses'' and ``holding sunglasses in hand'', to distinguish various cases. 
To erase each object for training, we use GroundingDINO~\cite{groundingdino} and SAM~\cite{sam} to obtain the object mask, and remove the object with an inpainting-based erasing model.
Specifically, we fine-tune an internal erasing model based on Flux.1 Fill~\cite{flux2024}.
Though without erasing capability, it is observed to quickly adapt to this task with a few training samples. 
To summarize, the pre-precessing pipeline outputs a set of triples, including the original image as $\mathcal{I}_T$, the object-erased image as $\mathcal{I}_P$, and the object textual description.

\textbf{Model Architecture: Text-to-Image \textit{v.s.} Inpainting Model.} 
There are two candidate variants of model to implement the mask-free try-on task, \textit{i.e.,} the text-to-image (T2I) model, and the mask-based inpainting model. 
Generally, mask-based VTON models~\cite{kim2024stableviton,chong2024catvton} leverage the fill-in capacity of inpainting model, while mask-free methods~\cite{zhang2024boow,DBLP:25Any2AnyTryon} adapt the T2I model, by injecting subject features into the backbone. 
Following the recent success in controllable image generation~\cite{omnicontrol,iclora,easycontrol}, a straightforward solution with T2I model is to concatenate the person image tokens into the sequence of noisy tokens, then processed with the full-attention mechanism in DiT. 
This strategy effectively transfers the person appearance into the target image, while also doubles the computation cost.

In contrast, \method explores to repurpose the inpainting model for mask-free generation. The inpainting model is generally finetuned from the T2I model via extending the input channels. Suppose the noisy latent as $X$, the input image as $I_c$, and the inpainting mask as $M$. Then the extended input is $\text{concat}(X;I_c (1-M); M)$, where $\text{concat}(\cdot)$ denotes channel-wise concatenation. 
For repurposing the model, we simply set $M=\mathbf{0}$, thus the input turns to be $\text{concat}(X;I_c;\mathbf{0})$. 
At the initialization, the zero mask leads the output image directly repeating the input.
Therefore, compared with T2I-based solution, the model effortlessly learns to copy the person condition, thus attentively focusing on locating the modification area.
We inject a location adapter (implemented as LoRA~\cite{lora}) for finetuning. 
In practice, the model converges rapidly to adapt the mask-free generative manner.

\begin{figure}[t]
    \centering
    \includegraphics[width=0.95 \textwidth]{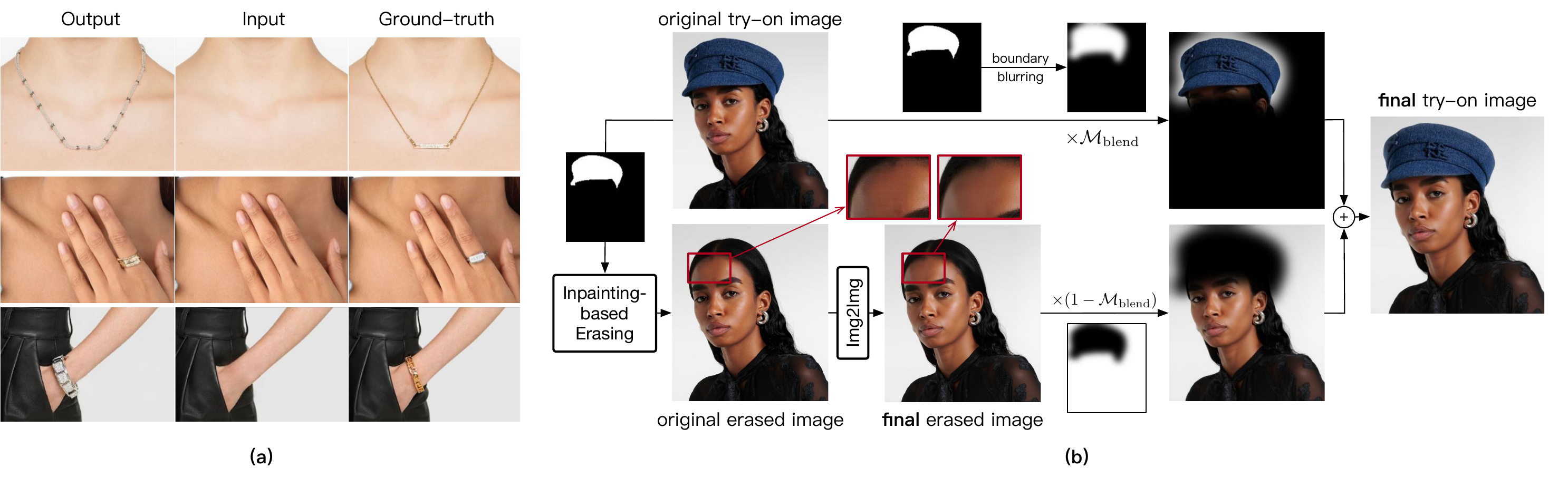}
    \vspace{-12pt}
     \caption{\textbf{Study on traceless erasing.} 
     (a) Shortcuts learned by model with naive erasing, where the model recovers the same shape and position as the ground-truth.
     (b) The pipeline of traceless erasing, where image-to-image model is introduced to disturb the traces (indicated in the red boxes).  
     }
    \label{fig:traceless_erasing}
\end{figure}

\textbf{Traceless Erasing.} 
Early experimental results suggest that the model learn unexpected shortcut.
We visualize the training monitoring result in \cref{fig:traceless_erasing} (a), where we evaluate the model on erased training samples. 
It is shown that the output image almost perfectly recovers the position and shape of the object in ground-truth image, which indicates information leakage.
We attribute the problem to the erasing model that leaves invisible traces in the filling area~\cite{wang2023dire,zhong2023rich}.
The model tends to figure out these abnormal area for editing, instead of predicting the reasonable position. When applying to real-world images, the model frequently fails to locate the try-on area, and directly repeat the input.

To address this problem, we propose a traceless erasing strategy, as shown in \cref{fig:traceless_erasing} (b). 
After erasing the object with inpainting model, we apply with an image-to-image (I2I) translation~\cite{meng2021sdedit} to subtly re-paint the image. We first add noise to the erased image $\hat{\mathcal{I}}_P$, referring a specific timestep $t \in [0,1]$ in diffusion schedule ($t=0.2$ in this paper), \textit{i.e.,} $z = enc(\hat{\mathcal{I}}_P)\times (1-t)+\epsilon \times t $, where $enc(\cdot)$ denotes the VAE encoder and $\epsilon$ is a standard Gaussian noise. 
Then a T2I model denoises $z$ into normal image with partial diffusion process from $t$ to $0$. 
In this manner, the artificial effects in inpainting area are confused with the whole re-painted image, thus avoid information leakage.
Since the I2I process modifies the detail of person image, the original try-on image should be correspondingly adjusted. 
To achieve smooth transition in the object boundary, we modulate the original mask $\mathcal{M}$ into a blending mask $\mathcal{M}_{blend}$ by blurring the boundary area for gradual blending effect.
The final try-on image is:
\begin{equation}
\vspace{-4pt}
    \mathcal{I}^\text{blend}_T = \mathcal{I}_T \times \mathcal{M}_\text{blend} + \text{img2img}(\hat{\mathcal{I}}_P) \times (1-\mathcal{M}_\text{blend})
    \vspace{-2pt}
\end{equation}

\subsection{Stage-2: ID Consistency Preservation}

The second stage of \method inherits the location adapter from stage one, and steps further to control the consistency of  object appearance. 
Referring to \cref{fig:pipeline}, in-shop image pairs are leveraged containing try-on image $\mathcal{I}_T$ and object image $\mathcal{I}_O$. 
We pre-process the data with traceless erasing, and gather a list of triple $(\mathcal{I}_T,\mathcal{I}_P,\mathcal{I}_O)$ for training. 
Considering the lack of enough samples, the objective is to conduct efficient training with minimal adjustment to the model architecture in stage one.

\textbf{Masked Full-Attention.} 
Following the recent full-attention customization researches~\cite{omnicontrol,iclora}, we directly append the object image tokens into the existing sequence in DiT, and shift their position embedding in the width dimension. Under this settings, OminiControl-2~\cite{tan2025ominicontrol2}
and EasyControl~\cite{easycontrol} also explore to block some information flow in attention. In detail, the attention mask is set to zero where the condition tokens serve as query and the generated tokens as key. Such an attention mask improves the inference efficiency, but leads to performance decrease to a certain extent. 

The main difference between the above works and \method is that the condition image is also concatenated with noisy latents and all-zero mask, for adaption to inpainting model. To cope with such variance, we design two strategies in training: (i) We compute diffusion loss on object image with itself as supervision, \textit{i.e.,} directly copying the input, which is aligned with the zero-mask input. (ii) We block all the data flow from the generating try-on image to object image, thus avoid the detailed object appearance to be interrupted. In practice, we find it helpful to better preserve object identity with the above masked full-attention.

\textbf{Two-Stream Adapters.} 
To fully preserve the ability of mask-free localization, we maintain the forward process of person image tokens exactly consistent with the first stage.
Then an identity adapter is initialized for the newly introduced object image tokens.
The two adapters, in same architecture, serve for a two-stream computation process, \textit{i.e.,} we switch different adapters by identifying tokens from different image sources.
The inference is similar to the multi-modality DiT~\cite{sd3} coping with vision and language information separately. 

\begin{figure}[tbp]
    \centering
        \centering
        \includegraphics[width=0.95\textwidth]{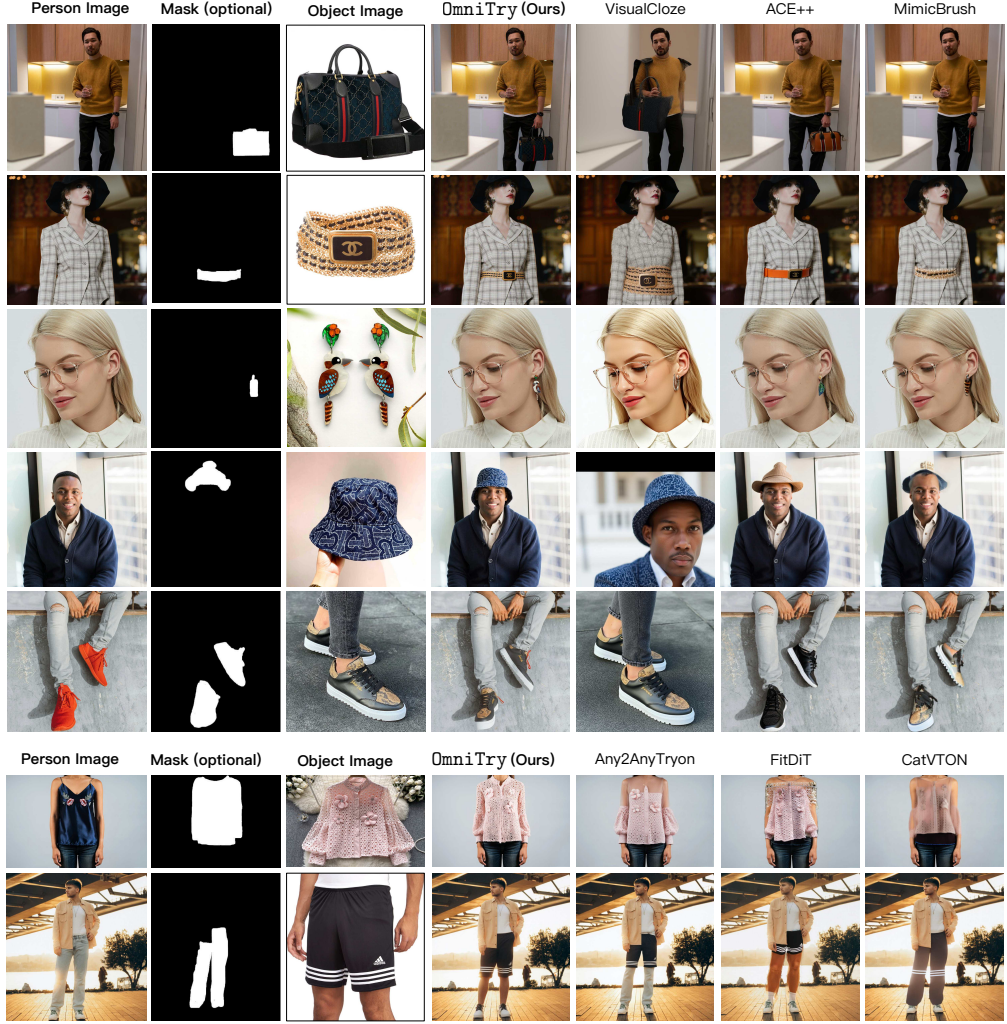}
        \vspace{-5pt}
        \caption{\textbf{Qualitative comparison} among \method and existing methods on multiple objects.
        }
        \vspace{-15pt}
        \label{fig:main_comp}
\end{figure}

\subsection{Evaluation Benchmark}
\label{chap:3_4_benchmark}
As the first work exploring unified virtual try-on task, we establish a comprehensive benchmark, dubbed \textit{OmniTry-Bench}, for better evaluation and comparison with existing works.

\textbf{Benchmark Collection.}
We gather evaluation samples within 12 common types of wearable objects, which can be summarized into 4 major classes: (i) \textit{clothes} consisting of top, bottom and full-body garments, (ii) \textit{shoes} in common styles, (iii) \textit{jewelries}, including bracelets, earrings, necklaces and rings, (iv) \textit{accessories}, including bags, belts, hats, glasses, sunglasses and ties.
We consider detailed sub-types if necessary, such as the backpack, shoulder and tote bags.
For each sub-type, we collect 15 paired test images for man and woman, separately. 
The object images are assigned in white background, natural background, and try-on setting,
with 5 pairs for each. 
The person images are also set in white and natural backgrounds.
Such settings ensure to fully evaluate the robustness of model.
Overall, the evaluation benchmark contains 360 pairs of images.

\textbf{Evaluation Metrics.}
As discussed in \cref{chap:3_1_preliminary}, the objectives of try-on can be divided into three aspects. Since there is no ground-truth result in mask-free setting, we redesign the metrics as follows:

\textit{Object Consistency}: We crop the objects from the try-on and object images via masking, and compute the visual similarity using DINO~\cite{dino} and CLIP~\cite{clip}, with metrics noted as M-DINO and M-CLIP-I.

\textit{Person Preservation}: In contrast, we crop out the person from try-on and person images, and compute spatial-aligned similarity between them, \textit{i.e.,} LPIPS~\cite{lpips} and SSIM~\cite{ssim}.

\textit{Object Localization}: 
(i) Counting the success rate whether a visual grounding model~\cite{groundingdino} detects the object, denoted as G-Accuracy. 
(ii) Computing the image-text similarity, noted as CLIP-I, between try-on image and a text describing the person trying on the object (generated by MLLM~\cite{bai2025qwen2}).

%% file: sections/4_experiment.tex
\section{Experiment}

\subsection{Experimental Setup}
\label{chap:5_1_exp_setup}

\textbf{Training Data.}
For the first stage, we gather a diverse dataset containing both in-the-wild portrait images and in-shop model shots. Considering each image could contain multiple wearable objects, the total amount of training pairs is $188,694$.
For the second stage, we collect paired samples following the 12 basic types in our benchmark. The whole dataset contains $51,195$ pairs, which shows class-unbalanced distribution ($14,861$ pairs for clothes and $295$ for ties). 
During training, each pair is equipped with a brief text description, such as ``trying on sunglasses'', to help distinguishing different classes. We note that the clothes and shoes are not erased but replaced with another one. Thus, we exchange the prefix as ``replacing'' for their prompts.

\textbf{Implementation Details.}
We train the first stage with batch-size of $32$ for $50$K steps, and the second stage with batch-size of $16$ for $25$K steps. All the experiments are conducted on $4$ NVIDIA H800 GPUs.
The location and identity adapters are implemented as LoRA~\cite{lora} with rank $16$. We employ the AdamW~\cite{adamw} optimizer with learning rate of $1^{-4}$ and weight decay of $0.01$. All the images are resized to a maximum of $1$ million pixels while preserving their original aspect ratios to training.

\textbf{Compared Methods.}
We primarily compare with methods in two basic paradigms:

\textit{Image-based Virtual Try-On}: 
Most VTON methods focus exclusively on garments. We compare on the clothes subset with representative works, including CatVTON~\cite{chong2024catvton}, OOTDiffusion~\cite{xu2025ootdiffusion}, Magic Clothing~\cite{DBLP:24MagicClothing}, FitDiT~\cite{DBLP:24FitDiT}, and Any2AnyTryon~\cite{DBLP:25Any2AnyTryon} (the only open-sourced mask-free model).


\textit{General Customized Image Generation}: 
Recent works explore to unify customization-related tasks into a single model, \textit{e.g.,} transferring the whole subject or local components, in mask-based or mask-free manners.
We compare with notable implementations, including Paint-by-Example~\cite{DBLP:23PaintbyExample}, MimicBrush~\cite{DBLP:24Mimicbrush}, ACE++~\cite{2025ACE}, OneDiffusion~\cite{DBLP:24OneDiffusion}, OmniGen~\cite{DBLP:24OmniGen} and VisualCloze~\cite{li2025visualclozeuniversalimagegeneration}.

To cope with the methods requiring masks of editing areas, we manually draw the masks in person image regarding the type of objects. Thus, the results of these methods are listed for reference, instead of direct comparison with the remaining mask-free methods.


\input{tables/main_exp}

\subsection{Results on Unified Virtual Try-on}

\textbf{Qualitative Results.}
We visualize the try-on examples generate by representative compared methods in \cref{fig:main_comp}.
For the general customization methods, mask-based works only modify the given areas, but show unstable object identity transferring. 
While for the mask-free works, the results tend to be a free combination of the input person and object. Though with better consistency, they fail to precisely preserve the person image. 
\method achieves accurate object consistency, in the meanwhile only edits the proper try-on areas of person image in mask-free manner. 
On the clothes subset evaluation, we observe that the existing VTON methods show unnatural output when evaluated on in-the-wild data.
\method is empowered by the compounded training on both in-the-wild and in-shop data, and shows more generalized ability of adapting various styles of garments.

\textbf{Quantitative Results.}
\cref{tab:main_exp} incorporates the evaluation results on the proposed OmniTry-Bench, conducted on the whole benchmark and the clothes subset, respectively.
\method outperforms existing methods on both sets.
For the mask-based customization methods, though the input mask helps to localize the editing area, they sometimes fail to transfer the complete appearance of objects, resulting in lower consistency metrics.
For the generalized customization methods in mask-free manner, they achieve better subject-ID preservation, but suffers to maintain the person image, thus show worse LPIPS and SSIM.
Such quantitative results are consistent with the visualized comparison results.
When evaluated on clothes subset, though \method is not specifically optimized on clothes dataset, it still shows advancing performance compared with state-of-the-art works in mask-based and mask-free settings. 
We note that the mask-free try-on could not be evaluated on previous benchmarks (\textit{e.g.,} VITON-HD~\cite{choi2021vitonhd}) for the missing of person images.

\begin{figure}[tbp]
    \centering
        \centering
        \includegraphics[width=\textwidth]{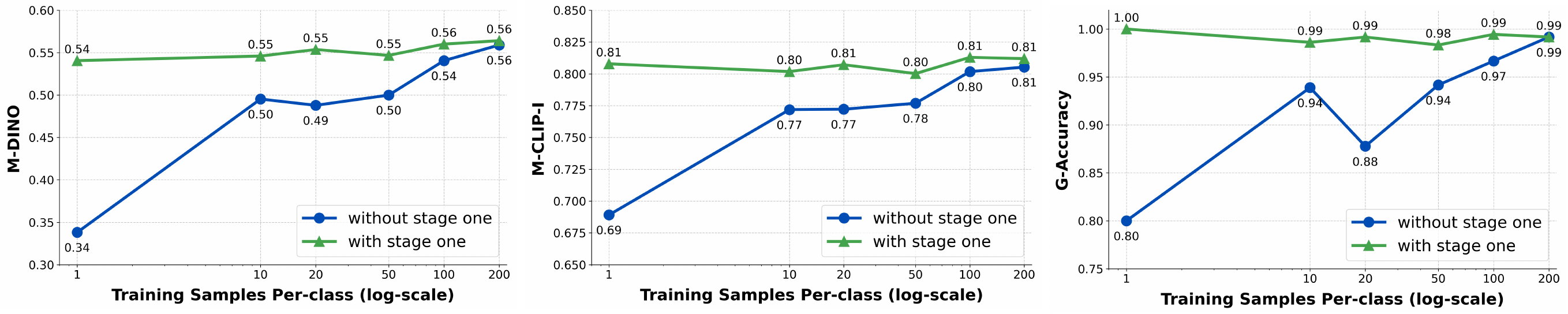}
        \vspace{-12pt}
        \caption{Ablation study on the two-staged training framework in \textbf{few-shot settings}. We show the evaluation metrics given varying amounts of paired training samples.}
        \label{fig:ab_train_samples}
\end{figure}

\subsection{Ablation Study}

\textbf{On the Training Strategy.}
We study one of the key designs in \method, \textit{i.e.,} the two-staged training framework. The first stage is intended to leverage large-scale unpaired data, and boost the training efficiency in the second stage.
To demonstrate this, we evaluate models initialized by the first stage and from scratch, respectively.
For the comparison of efficiency, the models are fine-tuned in few-shot settings, ranging from $1$ to $200$ training samples per class.
The results with representative metrics are illustrated in \cref{fig:ab_train_samples}. For metrics related to person preservation (LPIPS and SSIM), we note that they could be higher when the model fails and directly repeats the input, thus not included.

\input{tables/ablation}

It is observed that model from scratch shows increasing performance with more training samples per class. 
While for model initialized from the first stage, it already achieves satisfying performance even with only one example for training.
The results demonstrate that the first stage training significantly boosts the efficiency for fine-tuning, and is especially friendly to uncommon types of objects.
It is noted that though the few-shot tuning achieves good performance, we still fine-tune it with all available paired data to further increase the stability of model, referring to the results in \cref{tab:main_exp}

\textbf{On the Model Architecture.}
We then conduct ablation study on all the explored design of model architecture in \method. The results are shown in \cref{tab:ab_model}, where the ``full method'' indicates the final solution.
(i) We start with the comparison using text-to-image and inpainting model as backbone. Results show that the inpainting backbone performs better on all metrics which is consistent with our assumption that inpainting model takes no efforts to preserve the original image and converges faster.
(ii) For the additional loss computation on object image, we observe that removing the loss decreases the model performance to a certain extent.
(iii) For the attention mechanism, full attention additionally introduces flow from person to object image, thus the object consistency metrics decrease correspondingly.
(iv) We also investigate to use a single adapter for this task, \textit{i.e.} applying the adapter from the first stage to all image tokens. 
The one-stream framework also decreases the model performance, since it plays different roles in the inference of person and object images.

\textbf{On the Traceless Erasing.}
To verify the effectiveness of traceless erasing, we conduct ablation study on the jewelry subset with naive and traceless erasing.
Results in \cref{tab:ab_model} suggest that removing the traceless erasing leads to dramatic decrease in all metrics.
Therefore, we adopt traceless erasing as a fundamental pre-processing strategy in \method.

\subsection{Extension to Uncommon Classes}

\begin{figure}[tbp]
    \centering
        \centering
        \includegraphics[width=\textwidth]{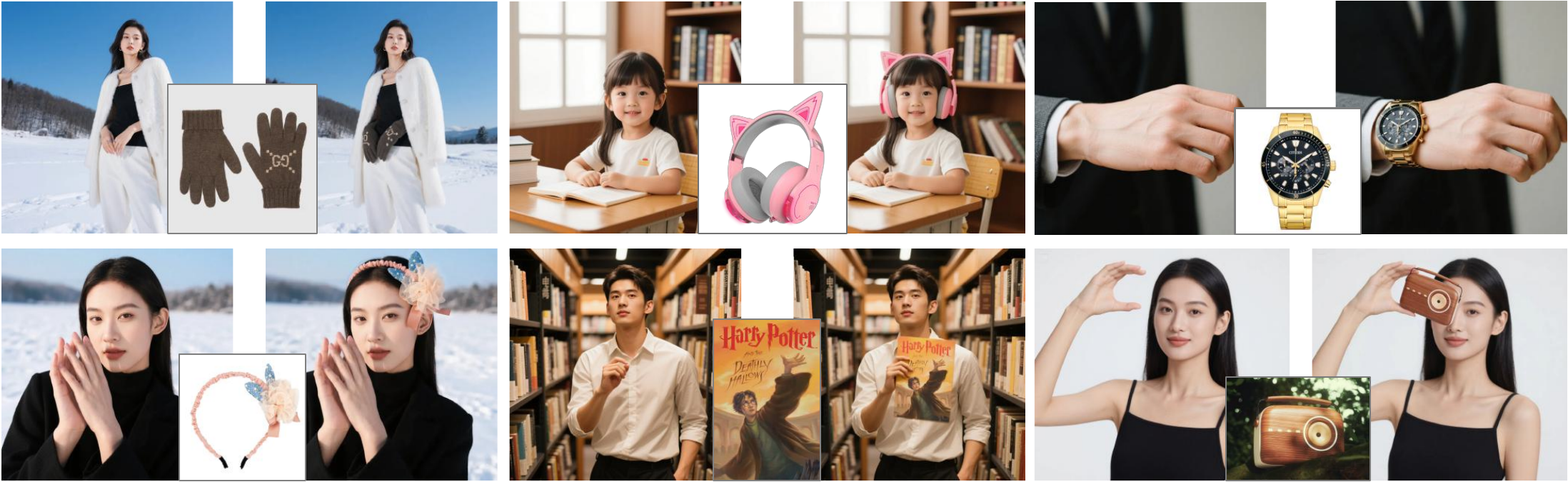}
        \vspace{-5pt}
        \caption{Try-on results of \method fine-tuned on uncommon classes of wearable or holdable objects.}
        \label{fig:uncommon_classes}
\end{figure}

We evaluate \method on 12 common types of objects in the main experiments.
To further demonstrate the efficiency of \method, we extend it to some uncommon types, for which the paired training samples are limited to be obtained.
The experiment is conducted on types including gloves, earphones, watches, hairbands, books and electronic products, with roughly $20$ samples per class.
It is noted that some types like books are actually in broader definition of try-on, \textit{i.e.,} holdable items.

The visualization results are shown in \cref{fig:uncommon_classes}. 
Thanks to the generalized training of the first stage, though with few paired samples, \method succeeds in transferring these relatively uncommon  objects onto the correct position.
The results encourage broader extension of \method into more application scenarios, without preparing a large amount of paired images. 

%% file: tables/main_exp.tex
\begin{table}[t]
\caption{\textbf{Evaluation results on OmniTry-Bench}, which is separated into two groups: results on the whole set and the clothes subset, for fair comparison with methods only optimized on clothes data.}
\label{tab:main_exp}
\centering
\small
\setlength\tabcolsep{2pt}
\begin{tabular}{lc@{\extracolsep{4pt}}cc@{\extracolsep{4pt}}cc@{\extracolsep{4pt}}cc}
\toprule
                 &     & \multicolumn{2}{c}{\textbf{Object Consistency}}                    & \multicolumn{2}{c}{\textbf{Person Presevation}}               & \multicolumn{2}{c}{\textbf{Object Localization}}                 \\ 
\addlinespace[3pt] \cline{3-4}\cline{5-6}\cline{7-8} \addlinespace[3pt] 
method           & mask & M-DINO $\uparrow$ & M-CLIP-I $\uparrow$ & LPIPS $\downarrow$ & SSIM$\uparrow$ & G-Acc. $\uparrow$ & CLIP-T $\uparrow$ \\
\midrule
\multicolumn{8}{c}{\textit{on the whole set}}      \\    \midrule
Paint-by-Example~\cite{DBLP:23PaintbyExample} &    \multirow{3}{*}{\ding{51}}                      &         0.4565                   &                0.7727              &           0.3903                &       0.8033                   &               0.9861             &           0.2804                 \\
MimicBrush~\cite{DBLP:24Mimicbrush} &      &        0.4693             &            0.7253                &          0.3033                    &          0.8575                 &         0.9250                                         &               0.2781             \\
ACE++~\cite{2025ACE}            &                          &            0.4565                &            0.7474                  &             0.4561              &           0.7519               &              0.9667              &              0.2791              \\
\midrule
OneDiffusion~\cite{DBLP:24OneDiffusion}     &       \multirow{4}{*}{\ding{55}}                   & 0.4731                     & 0.7749                       &   0.7001  &	0.5831&               \textbf{0.9972}         &           0.2309           \\
VisualCloze~\cite{li2025visualclozeuniversalimagegeneration}      &                          & 0.5292                     & 0.7782                       & 0.4471                    & 0.6190                    &   0.9639                     & 0.2524                     \\
OmniGen~\cite{DBLP:24OmniGen}          &                          & 0.5435                     & 0.7869                       &      0.6703 &	0.5965           &    0.9944                   &  0.2535      \\  
\method (\textbf{Ours})    &  & \textbf{0.6160}   & 	\textbf{0.8327}   &   \textbf{0.0542}	  & 
  \textbf{0.9333}     &   \textbf{0.9972}   &  \textbf{0.2831}    \\
\midrule
\midrule
\multicolumn{8}{c}{\textit{on the clothes subset}} \\
\midrule
Magic Clothing~\cite{DBLP:24MagicClothing}    &  \multirow{4}{*}{\ding{51}} & 0.5665 & 	0.7634	 & 	0.2761	 & 0.8786 & 1.0 &    0.2700 \\
CatVTON~\cite{chong2024catvton}    &  & 0.5744 & 	0.7906 & 	0.2084 & 	0.8828 & 1.0 &   0.2797 \\
OOTDiffusion~\cite{xu2025ootdiffusion}    &  & 0.5961 & 	0.8016 & 	0.2178 & 	0.8865	 & 1.0 &  0.2761 \\
FitDiT~\cite{DBLP:24FitDiT}    &  & 0.6733 & 	0.8340 & 	0.1618 & 	0.9027 & 1.0 & 	0.2831\\
\midrule
Any2AnyTryon~\cite{DBLP:25Any2AnyTryon}   &  \multirow{2}{*}{\ding{55}}   & 0.6747	 & 
 0.8537	 & 	0.2089 & 	0.8969 & 1.0 &  \textbf{0.2832} \\
\method (\textbf{Ours})   &  & \textbf{0.6995}   & 	\textbf{0.8560}  & 	\textbf{0.1021}	  & \textbf{0.9105} 
 & \textbf{1.0} &  0.2799 \\
\bottomrule
\end{tabular}
\end{table}

%% file: tables/ablation.tex
\begin{wraptable}{r}{0.5\textwidth}
\vspace{-5mm}  
    \tablestyle{5pt}{1.0}
    \def\w{20pt} 
    \setlength\tabcolsep{2pt}
    \caption{Ablation study on the model architecture and erasing strategies of \method.}
    \resizebox{\linewidth}{!}{
      \begin{tabular}{lcccc}
        \toprule
        method & M-DINO $\uparrow$ & M-CLIP-I $\uparrow$ & LPIPS $\downarrow$ & CLIP-T $\uparrow$\\
        \midrule
          \multicolumn{5}{c}{\textit{on the model architecutre (the whole subset)}} \\
        \midrule
        Full Method & \textbf{0.5991}& \textbf{0.8272} & 0.0557 & 0.2830\\
        - txt2img model & 0.5005&  0.7727& 0.0676& 0.2767\\
        - w.o. object loss & 0.5851&  0.8222& 0.0420& 0.2824\\
        - full attention & 0.5752& 0.8130& \textbf{0.0384} & \textbf{0.2832}\\
        - one-stream adapter & 0.5840& 0.8186& 0.0502& 0.2802\\
        \midrule
        \midrule
        \multicolumn{5}{c}{\textit{on the erasing strategy (the jewelry subset)}} \\
        \midrule
        naive erasing & 0.4964 & 0.7554 & 0.0413 & 0.2727 \\
        traceless erasing & \textbf{0.5389} & \textbf{0.7782} & \textbf{0.0288} & \textbf{0.2732} \\
         \bottomrule
    \end{tabular}
    }
    \label{tab:ab_model}
    \vspace{-4mm}  
\end{wraptable}
%


%% file: sections/5_conclusion.tex
\section{Limitations}
\label{chap_limitation}
In this section, we discuss the limitations of \method observed in practice. As the first work exploring unified VTON, \method is still restricted by the object types in training dataset. For the efficient tuning in stage-2, it could be challenging to extend to uncommon objects not involved in the unpaired dataset in stage-1. Larger pre-training dataset is expected to further boost the generalization ability.
For the mainly-focused 12 common types, experimental results show that \method could also fail to transfer the object consistency or output poor appearance in some cases, especially for the objects with larger transformation, \textit{e.g.,} bags.
The above limitations encourage future works to build upon \method and develop more advanced models towards unified try-on task.

\section{Conclusion}

This paper presents \method, a unified mask-free framework extending the existing garment try-on into any wearable objects.
To tackle the problem of lacking abundant paired samples, \textit{i.e.,} object and the try-on image, for many types of objects, we propose a two-staged training pipeline in \method.
During the first stage, large-scale unpaired images are leveraged to supervise the model for mask-free object localization.
While the second stage tames the model to maintain the object consistency.
We elaborate the design of \method, including a traceless erasing for avoiding shortcut learning, an inpainting-based re-purposing strategy for mask-free generation, and a masked full-attention for identity transferring.
A new benchmark targeting unified try-on is introduced, and demonstrates the effectiveness of \method compared with existing methods.
Extensive experiments also verify that \method achieves efficient learning even with few paired images for training.

%% file: sections_appendix/benchmark_details.tex
\section{Details of Benchmark and Metrics}
As the pioneering work investigating the unified virtual try-on task, we construct a comprehensive evaluation benchmark named \textit{OmniTry-Bench}, accompanied by six dedicated metrics to systematically assess the quality of synthesized try-on images.  
\begin{figure}[hbp]
    \centering
        \centering
        \includegraphics[width=0.999\textwidth]{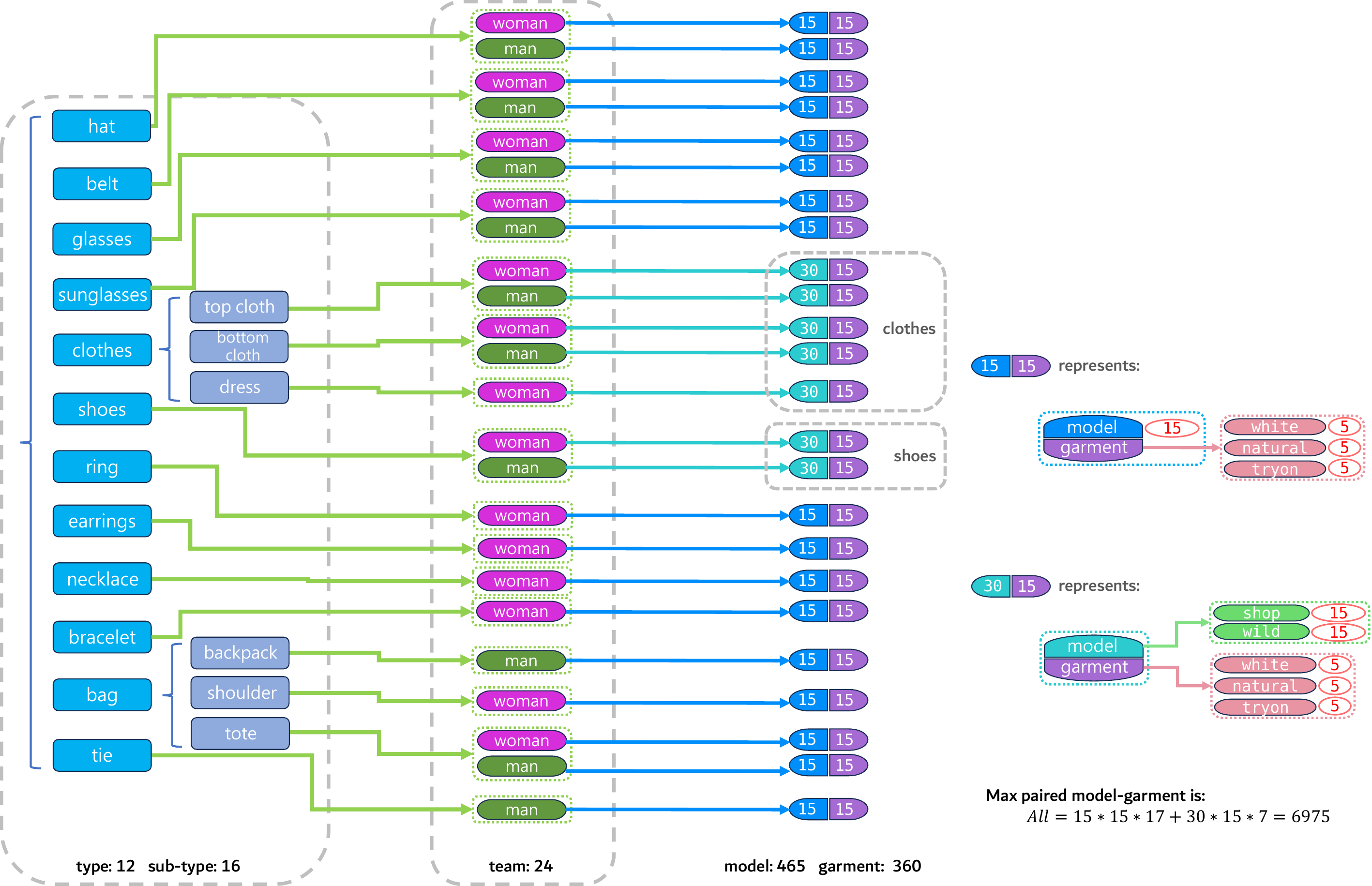}
        \vspace{-5pt}
        \caption{The visualization of the \textit{OmniTry-Bench} constitution.
        }
        \vspace{-1em}
        \label{fig:benchmark}
\end{figure}

\subsection{Constitution of Benchmark}

As the Figure~\ref{fig:benchmark}, we gather evaluation samples within 12 common types of wearable objects, which can be summarized into 4 major classes: (i) \textit{clothes} consisting of top, bottom and full-body garments, (ii) \textit{shoes} in common styles, (iii) \textit{jewelries}, including bracelets, earrings, necklaces and rings, (iv) \textit{accessories}, including bags, belts, hats, glasses, sunglasses and ties. 

We consider detailed sub-types if necessary, such as the class \textit{bag} consisted of the backpack, shoulder and tote bags.
\textit{Clothes} are divided into top cloth, bottom cloth, and dress. Each sub-type contains two gender groups (woman and man), with the exceptions that \textit{jewelries} and \textit{dress} exclusively contain woman samples, while \textit{tie} contains only man samples. 

Each gender group includes 15 model images, where the garments are categorized into three settings: white background, natural background, and try-on setting. Every garment setting include 5 images.
Following previous work's categorization of virtual try-on scenarios into \textit{in-shop} and \textit{in-the-wild}, we further divide the model images for \textit{clothes} and \textit{shoes} into 15 shop-style and 15 wild-style samples per gender group, resulting in 30 model images per sub-type.

The benchmark predominantly sources images from public repositories (Pexels\footnote{\url{https://www.pexels.com}}), supplemented with brand website materials and social media content under compliant data usage protocols.

\textbf{Pairing Strategy.} For each gender group, we establish combinatorial pairs between model and garment images through:
\begin{itemize}
\item \textit{Maximum Pair Calculation}:
$ max\_pairs = 15 \times 15 \times 17+30 \times 15 \times 7=6,975 $ pairs,
where 17 and 7 denote model settings counts for regular and style -specific categories respectively.
\item \textit{Sampled Pair Selection}:  
$ selected\_pairs = 15 \times 15 \times 24 = 360 $ paired samples,  
constrained by single-use garment policy and balanced sampling (15 models per clothes/shoes type, include 7 shop-style and 8 wild-style).
\end{itemize}

Overall, our experiments are all evaluated on the selected benchmark contains 360 pairs of images

\subsection{Evaluation Metrics}

As discussed before, the objectives of try-on can be divided into three aspects. Since there is no ground-truth result in mask-free setting, we redesign the metrics as follows:

\textit{Object Consistency}: We crop the objects from the try-on and object images via masking, then perform white-background normalization on the extracted objects. We compute the visual similarity using DINO~\cite{dino} and CLIP~\cite{clip} visual encoders, with metrics denoted as \text{M-DINO} and \text{M-CLIP-I}. As these metrics measure cosine similarity in the embedding space, their values range in $[-1, 1]$ where higher values indicate better object preservation.
The M-DINO scores generally exhibit lower values than M-CLIP-I, as DINO-extracted features are more sensitive to geometric variations compared to CLIP's semantic-aligned embeddings. Our experiments quantitatively validate this behavior across different object categories. This discrepancy stems from their distinct learning objectives:
\begin{itemize}
\item \textbf{M-DINO}~\cite{dino}: Learns dense local features through self-supervised distillation, emphasizing spatial consistency of object parts. Then compute the cosine similarity of two features.
\item \textbf{M-CLIP-I}~\cite{clip}: Optimizes global semantic alignment between object images, prioritizing category-level coherence. Then compute the cosine similarity of two features. Then compute the cosine similarity of two object features.
\end{itemize}

\textit{Person Preservation}: We extract the person regions by cropping try-on and original person images, masking the target object areas with black pixels. We then compute spatial-aligned similarity between these aligned image pairs using two complementary metrics:

\begin{itemize}
\item \textbf{SSIM} (Structural Similarity Index)~\cite{ssim}: Measures structural, luminance, and contrast similarity between images. The metric ranges in $[-1, 1]$ with values approaching 1 indicating higher structural consistency.
\item \textbf{LPIPS} (Learned Perceptual Image Patch Similarity)~\cite{lpips}: Computes deep feature differences using pretrained VGG networks, better aligning with human perception than traditional metrics. Its values lie in $[0, 1]$ where lower scores denote better preservation quality.
\end{itemize}

\textit{Object Localization}:
We propose a dual-strategy evaluation framework to assess spatial rationality through complementary approaches:

\begin{itemize}
\item \textbf{G-Accuracy}: Quantifies detection reliability using GroundingDINO~\cite{groundingdino} with the following implementation protocol:
Invoke \textit{predict\_with\_classes} API with target object categories as \textit{classes} parameter. Configure detection thresholds: $\textit{box\_threshold}=0.25 $ (bounding box confidence) and $\textit{text\_threshold}=0.25$ (text-image alignment). Last, calculate success rate as  total test cases correct detections.

\item \textbf{CLIP-I}: Evaluates semantic alignment through multi-modal similarity measurement: Generate descriptive prompts via Qwen2~\cite{bai2025qwen2} MLLM. Compute CLIP~\cite{clip} embedding similarity between try-on images and generated text. Normalize scores to [0,1] range using min-max scaling.

The final prompt template is formally defined as follows:

\begin{verbatim}
"""Generate a detailed description of a composite image by combining 
elements from the two provided images:
   1. Image 1: The model's appearance (pose, clothing, facial featur-
   es), background and style
   2. Image 2: Only the <{garment_class}>, without any other infos
   (e.g., background, model)
Describe the synthesized image with the model wearing the {garment_cl-
ass}, in 65 words. Only describe the final imagined scene, without the 
detail or information of composite. The main description is from 
Image 1. Briefly and shortly describe the {garment_class} in 6 wor-
ds, no details needed. No words like (e.g., from the Image 2). If 
{garment_class} is cloth or dress , the model from the Image 1, re-
place with the {garment_class} from Image 2, no words like (replace 
the hair/shirt), using "wear" the {garment_class}. 

Examples outputs:
    - "A young woman standing in a studio with a white background. 
    She is wearing a denim dress with a button-down collar and long 
    sleeves. The dress is knee-length and falls above her knees. T-
    he woman is also wearing black ankle boots with a pointed toe 
    and a low heel. She has a brown crossbody bag with a strap acr-
    oss her shoulder. The bag appears to be made of leather and has 
    a small flap closure. The overall style of the outfit is casual
    and minimalistic." 
    - "A close-up portrait of a young woman's face and upper body. 
    She is wearing a black strapless top with a thin silver chain n-
    ecklace around her neck. Her hair is styled in loose waves and 
    she is wearing large hoop earrings. The woman is looking off to 
    the side with a serious expression on her face. The background 
    is plain white."
    - "A close-up portrait of a woman's upper body. She is wearing 
    a black collared shirt with a button-down collar and long slee-
    ves. Her hair is styled in loose curls and she is wearing large, 
    dangling earrings. Her hand is resting on her chest, with a lar-
    ge ring on her ring finger. The background is plain white. The 
    woman appears to be looking off to the side with a serious expr-
    ession on her face."
"""

\end{verbatim}

\end{itemize}


%% file: sections_appendix/train_dataset_details.tex
\section{Details of Training Dataset}

\begin{wrapfigure}{r}{0.5\textwidth}
  \begin{center}
      \caption{The class distribution of training dataset.}
    \includegraphics[width=0.48 \textwidth]{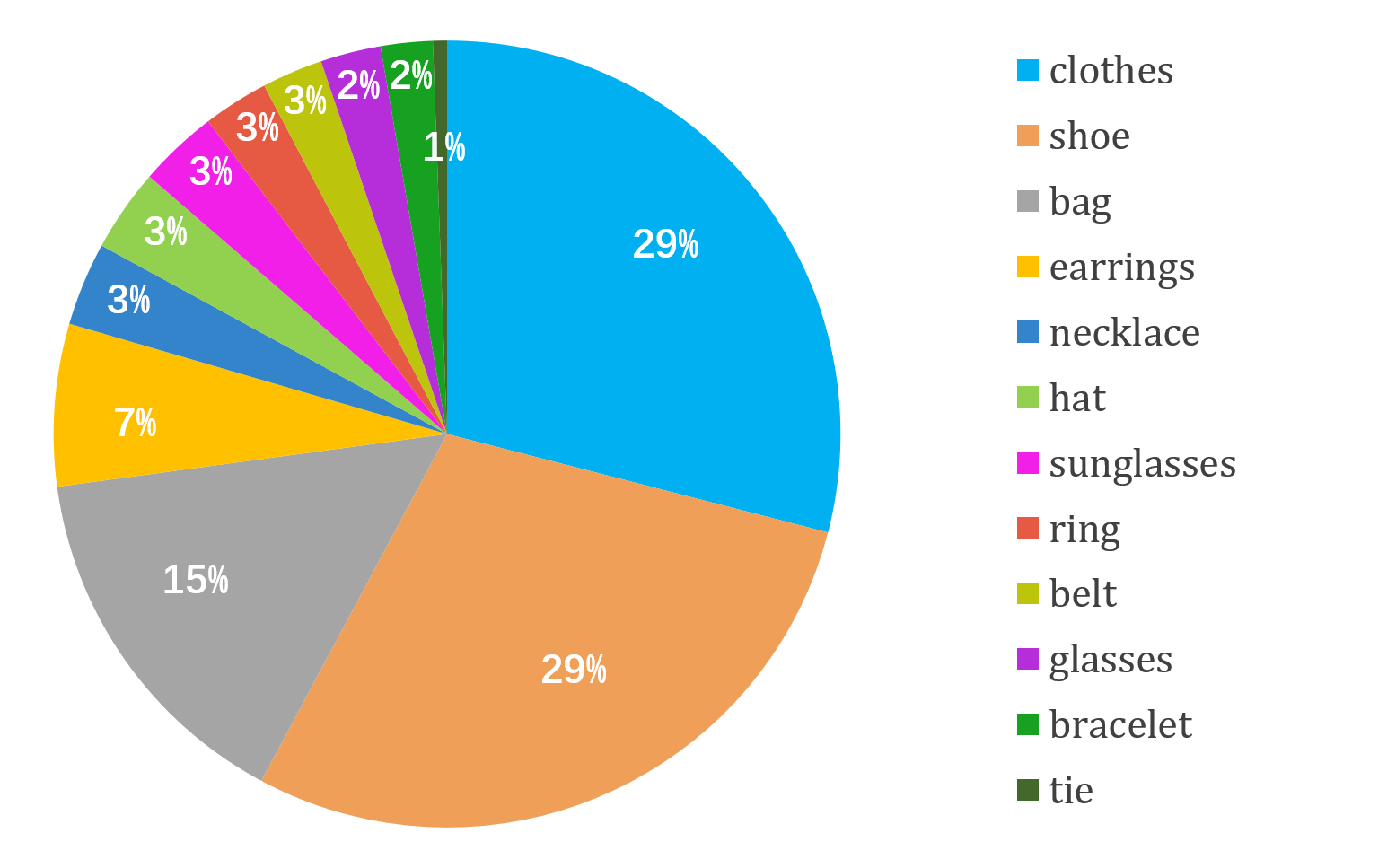}  \end{center}
  \label{fig:dataset_statistics}
\end{wrapfigure}

\subsection{Dataset for Stage-1}
The model in the first stage is jointly trained on two datasets, \textit{i.e.,} the unpaired in-the-wild images, and the dataset of stage-2 without the object image. We train on the datasets with sampling ratios of $2:1$. 
To further investigate the class distribution in the unpaired dataset, we count the highly-frequent words in the object text descriptions.
After filtering out the prepositions and verbs, the top-5 words are necklace, hat, glasses, sunglasses and watch. We also observe some classes excluded in our final 12 common classes, \textit{e.g.,} smartphone, cup, scarf, crown and mask. The rich distribution of wearable or holdable objects enhances the generalization of \method to uncommon classes.

We also report the scale of dataset during the data preparation. The initial dataset contains 152K in-the-wild images, which are filtered to be 111K images with person and wearable objects. After listing, grounding and removing objects, the total amount of images containing at least one object is 94K, and the corresponding number of objects is 189K (roughly 2 objects per image).

\subsection{Dataset for Stage-2} 
For the training dataset of the second stage, we visualize the amount of samples for each class in \cref{fig:dataset_statistics}
. It is shown that the most common classes, \textit{i.e.,} clothes and shoes, constitute more half of the total dataset, while most classes lay in the long-tail of distribution with less than $3\%$. 
Such a distribution is aligned with our basic assumption that it is hard to obtained paired samples for many wearable objects.
For class-balanced training, we manfully assign the sampling weights for clothes, shoes and bags as $4, 4, 3$, and set weights as $1$ for remaining classes.

%% file: sections_appendix/model_details.tex
\section{Details of Training and Model Architecture}

\subsection{Training Configuration}

During training, we resize the image with fixed aspect ratio to be no larger than $1$ million, which means that the model could receive images with varying aspect ratios in one batch. To handle this, we pad the image tokens into the same length of sequence, and modify the attention block to forward only on the valid tokens.

For both training of stage-1 and stage-2, we set the learning rate as  $1^{-4}$, gradient accumulation steps as $1$, weight decay as $0.01$ and gradient norm clipping as $1.0$. 
We use the AdamW~\cite{adamw} optimizer with hyper-parameters $\beta_1=0.9$ and $\beta_2 = 0.999$. 
The model is trained with mixed precision of bfloat16.
We note that since we fine-tune based on the distilled version of FLUX~\cite{flux2024}, the guidance scale is fixed as $1$ during training, and set as $30$ during inference.

\subsection{Details of Re-purposing Inpainting Model}
We elaborate the details of adapting the inpainting model, FLUX.1-Fill in this paper, towards mask-free try-on task.
During training, the input of model can be split into two sets in sequence dimension:
\begin{itemize}
    \item The try-on image. Along the channel dimension, it contains the noisy ground-truth try-on image, the input person image and a zero mask in the same shape.
    \item The object image. Along the channel dimension, it contains the noisy object image, the clean noisy image and a zero mask.
\end{itemize}

Then during the inference stage, we initialize the above input while replacing the noisy latents with standard Gaussian noise. 
Through the above formulation, it is shown that the inputs of person and object images are different. The person branch aims to modify the input person image in proper area, while the object branch simply targets to maintain the input, and transfers the object appearance via full attention mechanism.

\subsection{Details of Masked Full-Attention}
We discuss the details of applying masked full-attention in the second stage. We set text prompts for both try-on and object images, like ``trying on sunglasses''. Suppose the length of tokens to be: $L_{I1}$ for try-on image, $L_{T1}$ for try-on text, $L_{I2}$ for object image, and $L_{T2}$ for object text. We concatenate all tokens in the above order. Then the attention mask is:

\begin{equation}
    \begin{bmatrix}
        1_{L_{I1} \times L_{I1}} & 1_{L_{I1} \times L_{T1}} & 1_{L_{I1} \times L_{I2}} & 0_{L_{I1} \times L_{I1}}\\
        1_{L_{I1} \times L_{I1}} & 1_{L_{I1} \times L_{T1}} & 0_{L_{I1} \times L_{I2}} & 0_{L_{I1} \times L_{I1}}\\
        0_{L_{I1} \times L_{I1}} & 0_{L_{I1} \times L_{T1}} & 1_{L_{I1} \times L_{I2}} & 1_{L_{I1} \times L_{I1}}\\
        0_{L_{I1} \times L_{I1}} & 0_{L_{I1} \times L_{T1}} & 1_{L_{I1} \times L_{I2}} & 1_{L_{I1} \times L_{I1}}
    \end{bmatrix},
\end{equation}
where $1_{m\times n}$ denotes all-one matrix and $0_{m \times n}$ denotes all-zero matrix.
More specifically, we apply such a full-attention in both the multi-modality blocks and single blocks of FLUX~\cite{flux2024}, and figure out the text tokens to achieve the masking. We leverage the attention function with varying length in FlashAttention~\cite{dao2023flashattention2} to implement the block-wise masked attention.

\subsection{LoRA Implementation}
We implement the location and identity adapters with LoRA~\cite{lora}. In detail, we set the rank and $\alpha$ to be $16$.
We insert the LoRA module into the following layers: the projection into query/key/value, output projection of attention, the linear layers in feedforward block, the layer normalization layer, the input patch projection, and the final output projection.

%% file: sections_appendix/comp_methods.tex
\section{Details of Compared Methods}

In this section, we present the details of compared methods and our implementation of them on try-on task. We also report more results of the variants of each method, among which we only report the best result in main experiment.

\subsection{General Customized Image Generation}

\textbf{OneDiffusion}~\cite{DBLP:24OneDiffusion}: A large-scale diffusion framework supporting bidirectional image synthesis across tasks. We evaluated its performance on mask-free/mask-based try-on through instruction-based cases. We also modify its original instructing prompt to achieve better performance.

\textbf{OmniGen}~\cite{DBLP:24OmniGen}: A vision-language unified framework consolidating multiple tasks, supporting both mask-free/mask-based generation. We also test it with both standard and our optimized prompts.

\textbf{VisualCloze}~\cite{li2025visualclozeuniversalimagegeneration} implements visual in-context learning for domain generalization. We conduct experiments with single example and multiple examples in the context.

\textbf{Paint-by-Example}~\cite{DBLP:23PaintbyExample} enables to re-paint a given subject into image via CLIP-based object representation with mask dependency. 

\textbf{MimicBrush}~\cite{DBLP:24Mimicbrush} achieves imitative inpainting for region-specific edits, requiring the input image with mask, together with the reference image without mask.

\textbf{ACE++}~\cite{2025ACE} extends long-context conditioning for instruction-driven generation that tackles various.

\subsection{Image-based Virtual Try-On}

\textbf{OOTDiffusion}~\cite{xu2025ootdiffusion} designs a two-branch U-Net architecture to consume the person and garment images, which requires masked input in the person branch.

\textbf{Magic Clothing}~\cite{DBLP:24MagicClothing} introduces a garment extractor to progressively insert garment features into the main backbone of try-on generation. Magic Clothing supports the input of either masked person image, or the targeting pose and person ID image. We adapt the former setting to better preserve the person image.FI

\textbf{CatVTON}~\cite{chong2024catvton} proposes to transfer the identity of garment by simply concatenating it with the person image, and achieve mask-based try-on with inpainting model.

\textbf{FitDiT}~\cite{DBLP:24FitDiT} introduces diffusion transformer (DiT) model into VTON, and designs a GarmentDiT and a DenoisingDiT to implement this task.

\textbf{Any2AnyTryon}~\cite{DBLP:25Any2AnyTryon} is the only open-source mask-free VTON model, eliminates the dependence on masks, poses, or any other such conditions.

\subsection{More Comparison Results}

We report more comparison results in \cref{tab:more_comp_results}, including variants of methods with mask/mask-free setting, varying image size and different prompt design. We report only the best result of all variants in the main experiment.

\begin{table}[t]
\caption{More evaluation results of the compared methods with different settings.}
\label{tab:more_comp_results}
\centering
\footnotesize
\setlength\tabcolsep{2pt}
\begin{tabular}{lc@{\extracolsep{4pt}}cc@{\extracolsep{4pt}}cc@{\extracolsep{4pt}}cc}
\toprule
                 &     & \multicolumn{2}{c}{\textbf{Object Consistency}}                    & \multicolumn{2}{c}{\textbf{Person Presevation}}               & \multicolumn{2}{c}{\textbf{Object Localization}}                 \\ 
\addlinespace[3pt] \cline{3-4}\cline{5-6}\cline{7-8} \addlinespace[3pt] 
method           & mask & M-DINO $\uparrow$ & M-CLIP-I $\uparrow$ & LPIPS $\downarrow$ & SSIM$\uparrow$ & G-Acc. $\uparrow$ & CLIP-T $\uparrow$ \\
\midrule
\multicolumn{8}{c}{\textit{on the whole set}}      \\    \midrule
Paint-by-Example ($512^2$)~\cite{DBLP:23PaintbyExample} &    \multirow{10}{*}{\ding{51}}                      &         0.4171                   &                0.7328              &           0.4577                &       0.7968                   &               0.9833             &           0.2831                 \\
Paint-by-Example ($1024^2$)~\cite{DBLP:23PaintbyExample} &                      &         0.4565                   &                0.7727              &           0.3903                &       0.8033                   &               0.9861             &           0.2804                 \\
MimicBrush~\cite{DBLP:24Mimicbrush} &      &        0.4693             &            0.7253                &          0.3033                    &          0.8575                 &         0.9250                                         &               0.2781             \\
ACE++ (prompt v1)~\cite{2025ACE}            &                          &            0.4565                &            0.7474                  &             0.4561              &           0.7519               &              0.9667              &              0.2791              \\
ACE++ (prompt v2)~\cite{2025ACE}            &                          &            0.4449                &            0.7427                  &             0.4554              &           0.7517             &              0.9722              &              0.2793              \\
VisualCloze (1-example)~\cite{li2025visualclozeuniversalimagegeneration} & &	0.4705	& 0.7533 &	0.6685 &	0.5320 &  0.9972 & 0.2283 \\
VisualCloze (2-example)~\cite{li2025visualclozeuniversalimagegeneration} & &		0.4236	& 0.7307	&	0.6767 &	0.4908 & 0.9917 &  0.2260  \\
OmniGen (prompt v2)~\cite{DBLP:24OmniGen}  & & 0.5151 & 0.7761 & 0.6888 & 0.5870 & 0.9917 & 0.2557       \\
OneDiffusion (prompt v1)~\cite{DBLP:24OneDiffusion}   & &			0.5515 & 0.8137	&	0.6607 &	0.6166	& \textbf{1.0} & 0.2290	\\
OneDiffusion (prompt v2)~\cite{DBLP:24OneDiffusion}   & & 0.5580 &	0.7950	&	0.5795 &	0.6628 	& 0.9972 & 0.2401	\\

\midrule
OneDiffusion (prompt v1)~\cite{DBLP:24OneDiffusion}     &       \multirow{7}{*}{\ding{55}}                   & 0.4178                     & 0.7358                       &   0.7606  &	0.4951 &               \textbf{1.0}         &           0.2309           \\
OneDiffusion (prompt v2)~\cite{DBLP:24OneDiffusion}     &                      & 0.4731                     & 0.7749                       &   0.7001  &	0.5831 &               0.9972         &           0.2309           \\
VisualCloze (1-example)~\cite{li2025visualclozeuniversalimagegeneration}      &                          & 0.5292                     & 0.7782                       & 0.4471                    & 0.6190                    &   0.9639                     & 0.2524                     \\
VisualCloze (2-example)~\cite{li2025visualclozeuniversalimagegeneration}      &                          & 0.4915                     & 0.7619                       &      0.4730 &	0.5868           &    0.9806                   &  0.2540      \\  
OmniGen (prompt v1)~\cite{DBLP:24OmniGen}  & & 0.5299 & 0.7689 & 0.7009 & 0.5727 & 0.9778 & 0.2533       \\
OmniGen (prompt v2)~\cite{DBLP:24OmniGen} & & 0.5435                     & 0.7869                       &      0.6703 &	0.5965           &    0.9944                   &  0.2535      \\  
\method (\textbf{Ours})    &  & \textbf{0.6160}   & 	\textbf{0.8327}   &   \textbf{0.0542}	  & 
  \textbf{0.9333}     &   0.9972   &  \textbf{0.2831}    \\
\midrule
\midrule
\multicolumn{8}{c}{\textit{on the clothes subset}} \\
\midrule
Magic Clothing~\cite{DBLP:24MagicClothing}    &  \multirow{7}{*}{\ding{51}} & 0.5665 & 	0.7634	 & 	0.2761	 & 0.8786 & 1.0 &    0.2700 \\
CatVTON~\cite{chong2024catvton}    &  & 0.5744 & 	0.7906 & 	0.1664 & 	\textbf{0.9283} & 1.0 &   0.2818 \\
CatVTON (w. garment mask)~\cite{chong2024catvton}    &  & 0.5534 & 	0.7843 & 	0.2084 & 	0.8828 & 1.0 &   0.2797 \\
OOTDiffusion~\cite{xu2025ootdiffusion}    &  & 0.5961 & 	0.8016 & 	0.2178 & 	0.8865	 & 1.0 &  0.2761 \\
FitDiT ($768\times 1024$)~\cite{DBLP:24FitDiT}    &  & 0.6718 & 	0.8324 & 	0.1972 & 	0.8952 & 1.0 & 	0.2822\\
FitDiT ($1152\times 1536$)~\cite{DBLP:24FitDiT}    &  & 0.6733 & 	0.8340 & 	0.1618 & 	0.9027 & 1.0 & 	0.2831\\
FitDiT ($1536\times 2048$)~\cite{DBLP:24FitDiT}    &  & 0.5961 & 	0.8016 & 	0.2178 & 	0.8865 & 1.0 & 	0.2761\\
\midrule
Any2AnyTryon~\cite{DBLP:25Any2AnyTryon}   &  \multirow{2}{*}{\ding{55}}   & 0.6747	 & 
 0.8537	 & 	0.2089 & 	0.8969 & 1.0 &  \textbf{0.2832} \\
\method (\textbf{Ours})   &  & \textbf{0.6995}   & 	\textbf{0.8560}  & 	\textbf{0.1021}	  & 0.9105
 & \textbf{1.0} &  0.2799 \\
\bottomrule
\end{tabular}
\end{table}

%% file: sections_appendix/more_vis.tex
\section{More Visualization Results}

\begin{figure}[hbp]
    \centering
        \centering
        \vspace{-0.3em}
        \includegraphics[width=0.9999\textwidth]{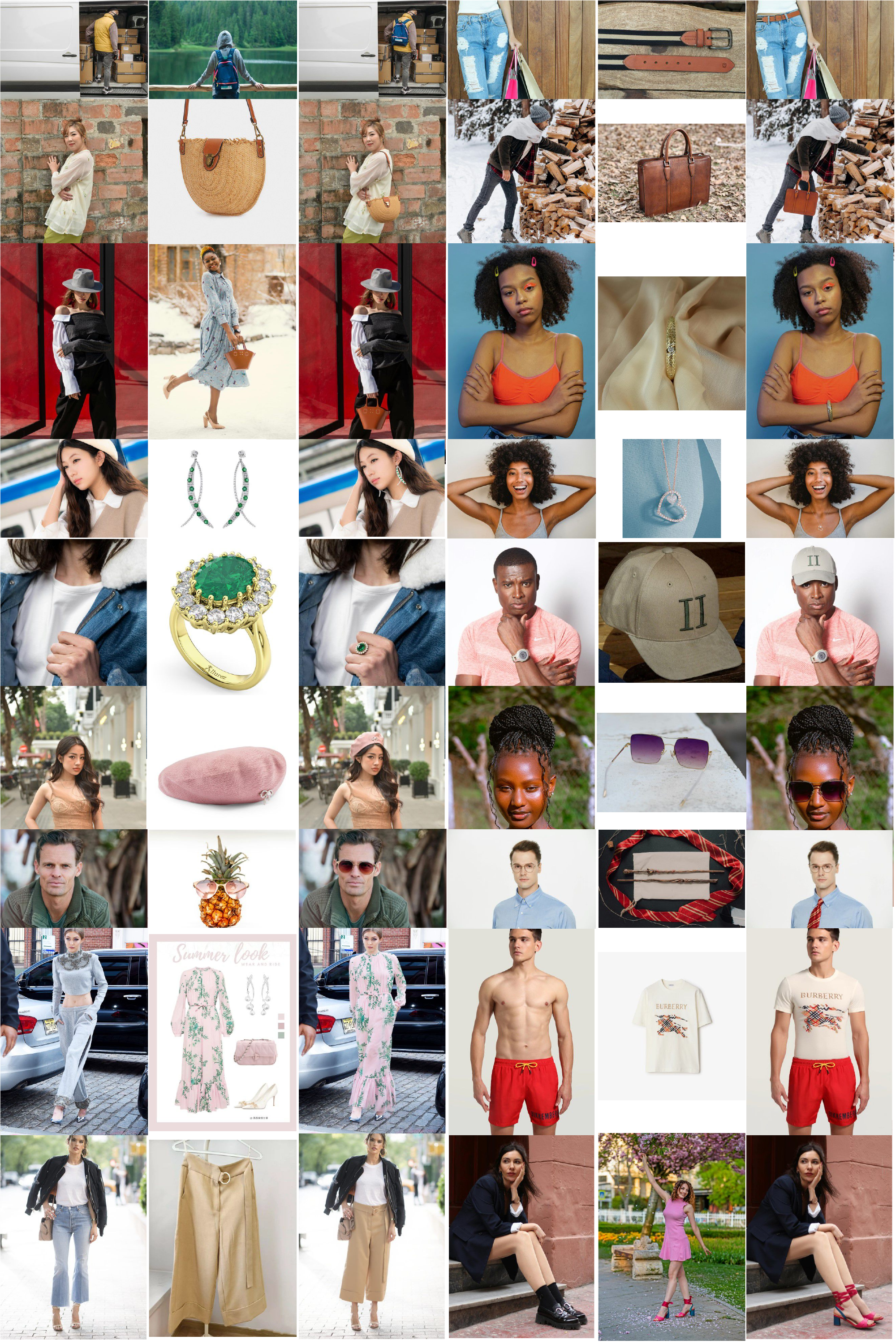}
        \vspace{-0.4em}
        \caption{The samples of the model, the object, and the try-on person.
        }
        \vspace{-1em}
        \label{fig:more_vis_results}
\end{figure}

We visualize more try-on results in \cref{fig:more_vis_results}, where we include all classes in OmniTry-Bench and different sub-types for full visualization.

%% file: arxiv.bbl
\begin{thebibliography}{70}
\providecommand{\natexlab}[1]{#1}
\providecommand{\url}[1]{\texttt{#1}}
\expandafter\ifx\csname urlstyle\endcsname\relax
  \providecommand{\doi}[1]{doi: #1}\else
  \providecommand{\doi}{doi: \begingroup \urlstyle{rm}\Url}\fi

\bibitem[Bai et~al.(2025)Bai, Chen, Liu, Wang, Ge, Song, Dang, Wang, Wang, Tang, et~al.]{bai2025qwen2}
Shuai Bai, Keqin Chen, Xuejing Liu, Jialin Wang, Wenbin Ge, Sibo Song, Kai Dang, Peng Wang, Shijie Wang, Jun Tang, et~al.
\newblock Qwen2. 5-vl technical report.
\newblock \emph{arXiv preprint arXiv:2502.13923}, 2025.

\bibitem[Bookstein(1989)]{DBLP:journals/pami/Bookstein89}
Fred~L. Bookstein.
\newblock Principal warps: Thin-plate splines and the decomposition of deformations.
\newblock \emph{{IEEE} Trans. Pattern Anal. Mach. Intell.}, 11\penalty0 (6), 1989.

\bibitem[Caron et~al.(2021)Caron, Touvron, Misra, J{\'e}gou, Mairal, Bojanowski, and Joulin]{dino}
Mathilde Caron, Hugo Touvron, Ishan Misra, Herv{\'e} J{\'e}gou, Julien Mairal, Piotr Bojanowski, and Armand Joulin.
\newblock Emerging properties in self-supervised vision transformers.
\newblock In \emph{Proceedings of the IEEE/CVF international conference on computer vision}, pages 9650--9660, 2021.

\bibitem[Chen et~al.(2024{\natexlab{a}})Chen, Gu, Xu, and Chen]{DBLP:24MagicClothing}
Weifeng Chen, Tao Gu, Yuhao Xu, and Arlene Chen.
\newblock Magic clothing: Controllable garment-driven image synthesis.
\newblock In \emph{Proceedings of the 32nd {ACM} International Conference on Multimedia, {MM} 2024, Melbourne, VIC, Australia, 28 October 2024 - 1 November 2024}. {ACM}, 2024{\natexlab{a}}.

\bibitem[Chen et~al.(2024{\natexlab{b}})Chen, Feng, Chen, Wang, Zhang, Liu, Shen, and Zhao]{DBLP:24Mimicbrush}
Xi~Chen, Yutong Feng, Mengting Chen, Yiyang Wang, Shilong Zhang, Yu~Liu, Yujun Shen, and Hengshuang Zhao.
\newblock Zero-shot image editing with reference imitation.
\newblock In \emph{Advances in Neural Information Processing Systems 38: Annual Conference on Neural Information Processing Systems 2024, NeurIPS 2024, Vancouver, BC, Canada, December 10 - 15, 2024}, 2024{\natexlab{b}}.

\bibitem[Chen et~al.(2024{\natexlab{c}})Chen, Huang, Liu, Shen, Zhao, and Zhao]{chen2024anydoor}
Xi~Chen, Lianghua Huang, Yu~Liu, Yujun Shen, Deli Zhao, and Hengshuang Zhao.
\newblock Anydoor: Zero-shot object-level image customization.
\newblock In \emph{Proceedings of the IEEE/CVF conference on computer vision and pattern recognition}, pages 6593--6602, 2024{\natexlab{c}}.

\bibitem[Chen et~al.(2024{\natexlab{d}})Chen, Zhang, Zhang, Zhou, Kim, Liu, Li, Zhang, Zhao, Wang, Ding, Lin, and Zhao]{DBLP:24UniReal}
Xi~Chen, Zhifei Zhang, He~Zhang, Yuqian Zhou, Soo~Ye Kim, Qing Liu, Yijun Li, Jianming Zhang, Nanxuan Zhao, Yilin Wang, Hui Ding, Zhe Lin, and Hengshuang Zhao.
\newblock Unireal: Universal image generation and editing via learning real-world dynamics.
\newblock \emph{CoRR}, 2024{\natexlab{d}}.

\bibitem[Choi et~al.(2021)Choi, Park, Lee, and Choo]{choi2021vitonhd}
Seunghwan Choi, Sunghyun Park, Minsoo Lee, and Jaegul Choo.
\newblock Viton-hd: High-resolution virtual try-on via misalignment-aware normalization.
\newblock In \emph{Proceedings of the IEEE/CVF conference on computer vision and pattern recognition}, pages 14131--14140, 2021.

\bibitem[Choi et~al.(2024)Choi, Kwak, Lee, Choi, and Shin]{idmvton}
Yisol Choi, Sangkyung Kwak, Kyungmin Lee, Hyungwon Choi, and Jinwoo Shin.
\newblock Improving diffusion models for virtual try-on.
\newblock \emph{arXiv e-prints}, pages arXiv--2403, 2024.

\bibitem[Chong et~al.(2024)Chong, Dong, Li, Zhang, Zhang, Zhang, Zhao, Jiang, and Liang]{chong2024catvton}
Zheng Chong, Xiao Dong, Haoxiang Li, Shiyue Zhang, Wenqing Zhang, Xujie Zhang, Hanqing Zhao, Dongmei Jiang, and Xiaodan Liang.
\newblock Catvton: Concatenation is all you need for virtual try-on with diffusion models.
\newblock \emph{arXiv preprint arXiv:2407.15886}, 2024.

\bibitem[Chou et~al.(2019)Chou, Lee, Zhang, Lee, and Hsu]{chou2019pivtons}
Chao-Te Chou, Cheng-Han Lee, Kaipeng Zhang, Hu-Cheng Lee, and Winston~H Hsu.
\newblock Pivtons: Pose invariant virtual try-on shoe with conditional image completion.
\newblock In \emph{Computer Vision--ACCV 2018: 14th Asian Conference on Computer Vision, Perth, Australia, December 2--6, 2018, Revised Selected Papers, Part VI 14}, pages 654--668. Springer, 2019.

\bibitem[Dao(2024)]{dao2023flashattention2}
Tri Dao.
\newblock Flash{A}ttention-2: Faster attention with better parallelism and work partitioning.
\newblock In \emph{International Conference on Learning Representations (ICLR)}, 2024.

\bibitem[Dosovitskiy et~al.(2020)Dosovitskiy, Beyer, Kolesnikov, Weissenborn, Zhai, Unterthiner, Dehghani, Minderer, Heigold, Gelly, et~al.]{vit}
Alexey Dosovitskiy, Lucas Beyer, Alexander Kolesnikov, Dirk Weissenborn, Xiaohua Zhai, Thomas Unterthiner, Mostafa Dehghani, Matthias Minderer, Georg Heigold, Sylvain Gelly, et~al.
\newblock An image is worth 16x16 words: Transformers for image recognition at scale.
\newblock \emph{arXiv preprint arXiv:2010.11929}, 2020.

\bibitem[Esser et~al.(2024)Esser, Kulal, Blattmann, Entezari, M{\"u}ller, Saini, Levi, Lorenz, Sauer, Boesel, et~al.]{sd3}
Patrick Esser, Sumith Kulal, Andreas Blattmann, Rahim Entezari, Jonas M{\"u}ller, Harry Saini, Yam Levi, Dominik Lorenz, Axel Sauer, Frederic Boesel, et~al.
\newblock Scaling rectified flow transformers for high-resolution image synthesis.
\newblock In \emph{Forty-first international conference on machine learning}, 2024.

\bibitem[Gal et~al.(2023)Gal, Alaluf, Atzmon, Patashnik, Bermano, Chechik, and Cohen{-}Or]{DBLP:23OneWord}
Rinon Gal, Yuval Alaluf, Yuval Atzmon, Or~Patashnik, Amit~Haim Bermano, Gal Chechik, and Daniel Cohen{-}Or.
\newblock An image is worth one word: Personalizing text-to-image generation using textual inversion.
\newblock In \emph{The Eleventh International Conference on Learning Representations, {ICLR} 2023, Kigali, Rwanda, May 1-5, 2023}, 2023.

\bibitem[Ge et~al.(2021)Ge, Song, Zhang, Ge, Liu, and Luo]{ge2021parser}
Yuying Ge, Yibing Song, Ruimao Zhang, Chongjian Ge, Wei Liu, and Ping Luo.
\newblock Parser-free virtual try-on via distilling appearance flows.
\newblock In \emph{Proceedings of the IEEE/CVF conference on computer vision and pattern recognition}, pages 8485--8493, 2021.

\bibitem[Gou et~al.(2023)Gou, Sun, Zhang, Si, Qian, and Zhang]{DBLP:23Taming}
Junhong Gou, Siyu Sun, Jianfu Zhang, Jianlou Si, Chen Qian, and Liqing Zhang.
\newblock Taming the power of diffusion models for high-quality virtual try-on with appearance flow.
\newblock In \emph{Proceedings of the 31st {ACM} International Conference on Multimedia, {MM} 2023, Ottawa, ON, Canada, 29 October 2023- 3 November 2023}, 2023.

\bibitem[Guo et~al.(2025)Guo, Zeng, Song, Zhang, Zhang, and Liu]{DBLP:25Any2AnyTryon}
Hailong Guo, Bohan Zeng, Yiren Song, Wentao Zhang, Chuang Zhang, and Jiaming Liu.
\newblock Any2anytryon: Leveraging adaptive position embeddings for versatile virtual clothing tasks.
\newblock \emph{CoRR}, 2025.

\bibitem[Han et~al.(2018)Han, Wu, Wu, Yu, and Davis]{han2018viton}
Xintong Han, Zuxuan Wu, Zhe Wu, Ruichi Yu, and Larry~S Davis.
\newblock Viton: An image-based virtual try-on network.
\newblock In \emph{Proceedings of the IEEE conference on computer vision and pattern recognition}, pages 7543--7552, 2018.

\bibitem[Ho et~al.(2020)Ho, Jain, and Abbeel]{ddpm}
Jonathan Ho, Ajay Jain, and Pieter Abbeel.
\newblock Denoising diffusion probabilistic models.
\newblock \emph{Advances in neural information processing systems}, 33:\penalty0 6840--6851, 2020.

\bibitem[Hu et~al.(2022)Hu, Shen, Wallis, Allen-Zhu, Li, Wang, Wang, Chen, et~al.]{lora}
Edward~J Hu, Yelong Shen, Phillip Wallis, Zeyuan Allen-Zhu, Yuanzhi Li, Shean Wang, Lu~Wang, Weizhu Chen, et~al.
\newblock Lora: Low-rank adaptation of large language models.
\newblock \emph{ICLR}, 1\penalty0 (2):\penalty0 3, 2022.

\bibitem[Hua et~al.(2023)Hua, Liu, Ding, Liu, Wu, and He]{DBLP:2023DreamTuner}
Miao Hua, Jiawei Liu, Fei Ding, Wei Liu, Jie Wu, and Qian He.
\newblock Dreamtuner: Single image is enough for subject-driven generation.
\newblock \emph{CoRR}, 2023.

\bibitem[Huang et~al.(2024)Huang, Wang, Wu, Shi, Dou, Liang, Feng, Liu, and Zhou]{iclora}
Lianghua Huang, Wei Wang, Zhi-Fan Wu, Yupeng Shi, Huanzhang Dou, Chen Liang, Yutong Feng, Yu~Liu, and Jingren Zhou.
\newblock In-context lora for diffusion transformers.
\newblock \emph{arXiv preprint arXiv:2410.23775}, 2024.

\bibitem[Issenhuth et~al.(2020)Issenhuth, Mary, and Calauzenes]{issenhuth2020not}
Thibaut Issenhuth, J{\'e}r{\'e}mie Mary, and Cl{\'e}ment Calauzenes.
\newblock Do not mask what you do not need to mask: a parser-free virtual try-on.
\newblock In \emph{Computer Vision--ECCV 2020: 16th European Conference, Glasgow, UK, August 23--28, 2020, Proceedings, Part XX 16}, pages 619--635. Springer, 2020.

\bibitem[Jiang et~al.(2024)Jiang, Hu, Luo, He, Xu, Peng, Zhang, Wang, Wu, and Fu]{DBLP:24FitDiT}
Boyuan Jiang, Xiaobin Hu, Donghao Luo, Qingdong He, Chengming Xu, Jinlong Peng, Jiangning Zhang, Chengjie Wang, Yunsheng Wu, and Yanwei Fu.
\newblock Fitdit: Advancing the authentic garment details for high-fidelity virtual try-on.
\newblock \emph{CoRR}, 2024.

\bibitem[Jiang et~al.(2025)Jiang, Wang, Bao, Zhou, Chen, Shi, Chen, and Li]{smarteraser}
Longtao Jiang, Zhendong Wang, Jianmin Bao, Wengang Zhou, Dongdong Chen, Lei Shi, Dong Chen, and Houqiang Li.
\newblock Smarteraser: Remove anything from images using masked-region guidance.
\newblock \emph{arXiv preprint arXiv:2501.08279}, 2025.

\bibitem[Kim et~al.(2024)Kim, Gu, Park, Park, and Choo]{kim2024stableviton}
Jeongho Kim, Guojung Gu, Minho Park, Sunghyun Park, and Jaegul Choo.
\newblock Stableviton: Learning semantic correspondence with latent diffusion model for virtual try-on.
\newblock In \emph{Proceedings of the IEEE/CVF conference on computer vision and pattern recognition}, pages 8176--8185, 2024.

\bibitem[Kingma et~al.(2013)Kingma, Welling, et~al.]{vae}
Diederik~P Kingma, Max Welling, et~al.
\newblock Auto-encoding variational bayes, 2013.

\bibitem[Kirillov et~al.(2023)Kirillov, Mintun, Ravi, Mao, Rolland, Gustafson, Xiao, Whitehead, Berg, Lo, et~al.]{sam}
Alexander Kirillov, Eric Mintun, Nikhila Ravi, Hanzi Mao, Chloe Rolland, Laura Gustafson, Tete Xiao, Spencer Whitehead, Alexander~C Berg, Wan-Yen Lo, et~al.
\newblock Segment anything.
\newblock In \emph{Proceedings of the IEEE/CVF international conference on computer vision}, pages 4015--4026, 2023.

\bibitem[Kumari et~al.(2023)Kumari, Zhang, Zhang, Shechtman, and Zhu]{DBLP:23MultiConcept}
Nupur Kumari, Bingliang Zhang, Richard Zhang, Eli Shechtman, and Jun{-}Yan Zhu.
\newblock Multi-concept customization of text-to-image diffusion.
\newblock In \emph{{IEEE/CVF} Conference on Computer Vision and Pattern Recognition, {CVPR} 2023, Vancouver, BC, Canada, June 17-24, 2023}, 2023.

\bibitem[Labs(2024)]{flux2024}
Black~Forest Labs.
\newblock Flux.
\newblock \url{https://github.com/black-forest-labs/flux}, 2024.

\bibitem[Le et~al.(2024)Le, Pham, Lee, Clark, Kembhavi, Mandt, Krishna, and Lu]{DBLP:24OneDiffusion}
Duong~H. Le, Tuan Pham, Sangho Lee, Christopher Clark, Aniruddha Kembhavi, Stephan Mandt, Ranjay Krishna, and Jiasen Lu.
\newblock One diffusion to generate them all.
\newblock \emph{CoRR}, 2024.

\bibitem[Li et~al.(2023)Li, Li, and Hoi]{DBLP:23BLIPdiffusion}
Dongxu Li, Junnan Li, and Steven C.~H. Hoi.
\newblock Blip-diffusion: Pre-trained subject representation for controllable text-to-image generation and editing.
\newblock In \emph{Advances in Neural Information Processing Systems 36: Annual Conference on Neural Information Processing Systems 2023, NeurIPS 2023, New Orleans, LA, USA, December 10 - 16, 2023}, 2023.

\bibitem[Li et~al.(2025)Li, Du, Yan, Zhuo, Li, Gao, Ma, and Cheng]{li2025visualclozeuniversalimagegeneration}
Zhong-Yu Li, Ruoyi Du, Juncheng Yan, Le~Zhuo, Zhen Li, Peng Gao, Zhanyu Ma, and Ming-Ming Cheng.
\newblock Visualcloze: A universal image generation framework via visual in-context learning, 2025.

\bibitem[Liu et~al.(2024)Liu, Zeng, Ren, Li, Zhang, Yang, Jiang, Li, Yang, Su, et~al.]{groundingdino}
Shilong Liu, Zhaoyang Zeng, Tianhe Ren, Feng Li, Hao Zhang, Jie Yang, Qing Jiang, Chunyuan Li, Jianwei Yang, Hang Su, et~al.
\newblock Grounding dino: Marrying dino with grounded pre-training for open-set object detection.
\newblock In \emph{European Conference on Computer Vision}, pages 38--55. Springer, 2024.

\bibitem[Liu et~al.(2022)Liu, Gong, and Liu]{liu2022flow}
Xingchao Liu, Chengyue Gong, and Qiang Liu.
\newblock Flow straight and fast: Learning to generate and transfer data with rectified flow.
\newblock \emph{arXiv preprint arXiv:2209.03003}, 2022.

\bibitem[Loshchilov and Hutter(2017)]{adamw}
Ilya Loshchilov and Frank Hutter.
\newblock Decoupled weight decay regularization.
\newblock \emph{arXiv preprint arXiv:1711.05101}, 2017.

\bibitem[Ma et~al.(2024)Ma, Liang, Chen, and Lu]{DBLP:24SubjectDiffusion}
Jian Ma, Junhao Liang, Chen Chen, and Haonan Lu.
\newblock Subject-diffusion: Open domain personalized text-to-image generation without test-time fine-tuning.
\newblock In \emph{{ACM} {SIGGRAPH} 2024 Conference Papers, {SIGGRAPH} 2024, Denver, CO, USA, 27 July 2024- 1 August 2024}, 2024.

\bibitem[Mao et~al.(2025)Mao, Zhang, Pan, Jiang, Han, Liu, and Zhou]{2025ACE}
Chaojie Mao, Jingfeng Zhang, Yulin Pan, Zeyinzi Jiang, Zhen Han, Yu~Liu, and Jingren Zhou.
\newblock Ace++: Instruction-based image creation and editing via context-aware content filling.
\newblock \emph{CoRR}, 2025.

\bibitem[Meng et~al.(2021)Meng, He, Song, Song, Wu, Zhu, and Ermon]{meng2021sdedit}
Chenlin Meng, Yutong He, Yang Song, Jiaming Song, Jiajun Wu, Jun-Yan Zhu, and Stefano Ermon.
\newblock Sdedit: Guided image synthesis and editing with stochastic differential equations.
\newblock \emph{arXiv preprint arXiv:2108.01073}, 2021.

\bibitem[Miao et~al.(2025)Miao, Huang, Han, Wang, Lin, and Shen]{miao2025shining}
Yingmao Miao, Zhanpeng Huang, Rui Han, Zibin Wang, Chenhao Lin, and Chao Shen.
\newblock Shining yourself: High-fidelity ornaments virtual try-on with diffusion model.
\newblock \emph{arXiv preprint arXiv:2503.16065}, 2025.

\bibitem[Morelli et~al.(2022)Morelli, Fincato, Cornia, Landi, Cesari, and Cucchiara]{morelli2022dresscode}
Davide Morelli, Matteo Fincato, Marcella Cornia, Federico Landi, Fabio Cesari, and Rita Cucchiara.
\newblock Dress code: High-resolution multi-category virtual try-on.
\newblock In \emph{Proceedings of the IEEE/CVF conference on computer vision and pattern recognition}, pages 2231--2235, 2022.

\bibitem[Morelli et~al.(2023)Morelli, Baldrati, Cartella, Cornia, Bertini, and Cucchiara]{DBLP:23LaDIvton}
Davide Morelli, Alberto Baldrati, Giuseppe Cartella, Marcella Cornia, Marco Bertini, and Rita Cucchiara.
\newblock Ladi-vton: Latent diffusion textual-inversion enhanced virtual try-on.
\newblock In \emph{Proceedings of the 31st {ACM} International Conference on Multimedia, {MM} 2023, Ottawa, ON, Canada, 29 October 2023- 3 November 2023}, 2023.

\bibitem[Mou et~al.(2024)Mou, Wang, Xie, Wu, Zhang, Qi, and Shan]{DBLP:24t2iAdapter}
Chong Mou, Xintao Wang, Liangbin Xie, Yanze Wu, Jian Zhang, Zhongang Qi, and Ying Shan.
\newblock T2i-adapter: Learning adapters to dig out more controllable ability for text-to-image diffusion models.
\newblock In \emph{Thirty-Eighth {AAAI} Conference on Artificial Intelligence, {AAAI} 2024, Thirty-Sixth Conference on Innovative Applications of Artificial Intelligence, {IAAI} 2024, Fourteenth Symposium on Educational Advances in Artificial Intelligence, {EAAI} 2014, February 20-27, 2024, Vancouver, Canada}, 2024.

\bibitem[Peebles and Xie(2023)]{dit}
William Peebles and Saining Xie.
\newblock Scalable diffusion models with transformers.
\newblock In \emph{Proceedings of the IEEE/CVF international conference on computer vision}, pages 4195--4205, 2023.

\bibitem[Podell et~al.(2023)Podell, English, Lacey, Blattmann, Dockhorn, M{\"u}ller, Penna, and Rombach]{sdxl}
Dustin Podell, Zion English, Kyle Lacey, Andreas Blattmann, Tim Dockhorn, Jonas M{\"u}ller, Joe Penna, and Robin Rombach.
\newblock {SDXL: Improving latent diffusion models for high-resolution image synthesis}.
\newblock \emph{arXiv preprint arXiv:2307.01952}, 2023.

\bibitem[Qin et~al.(2023)Qin, Zhang, Yu, Feng, Yang, Zhou, Wang, Niebles, Xiong, Savarese, Ermon, Fu, and Xu]{DBLP:23UniControl}
Can Qin, Shu Zhang, Ning Yu, Yihao Feng, Xinyi Yang, Yingbo Zhou, Huan Wang, Juan~Carlos Niebles, Caiming Xiong, Silvio Savarese, Stefano Ermon, Yun Fu, and Ran Xu.
\newblock Unicontrol: {A} unified diffusion model for controllable visual generation in the wild.
\newblock In \emph{Advances in Neural Information Processing Systems 36: Annual Conference on Neural Information Processing Systems 2023, NeurIPS 2023, New Orleans, LA, USA, December 10 - 16, 2023}, 2023.

\bibitem[Radford et~al.(2021)Radford, Kim, Hallacy, Ramesh, Goh, Agarwal, Sastry, Askell, Mishkin, Clark, et~al.]{clip}
Alec Radford, Jong~Wook Kim, Chris Hallacy, Aditya Ramesh, Gabriel Goh, Sandhini Agarwal, Girish Sastry, Amanda Askell, Pamela Mishkin, Jack Clark, et~al.
\newblock Learning transferable visual models from natural language supervision.
\newblock In \emph{International conference on machine learning}, pages 8748--8763. PmLR, 2021.

\bibitem[Rombach et~al.(2022)Rombach, Blattmann, Lorenz, Esser, and Ommer]{stablediffusion}
Robin Rombach, Andreas Blattmann, Dominik Lorenz, Patrick Esser, and Bj{\"o}rn Ommer.
\newblock High-resolution image synthesis with latent diffusion models.
\newblock In \emph{Proceedings of the IEEE/CVF conference on computer vision and pattern recognition}, pages 10684--10695, 2022.

\bibitem[Su et~al.(2024)Su, Ahmed, Lu, Pan, Bo, and Liu]{rope}
Jianlin Su, Murtadha Ahmed, Yu~Lu, Shengfeng Pan, Wen Bo, and Yunfeng Liu.
\newblock Roformer: Enhanced transformer with rotary position embedding.
\newblock \emph{Neurocomputing}, 568:\penalty0 127063, 2024.

\bibitem[Suvorov et~al.(2022)Suvorov, Logacheva, Mashikhin, Remizova, Ashukha, Silvestrov, Kong, Goka, Park, and Lempitsky]{lama}
Roman Suvorov, Elizaveta Logacheva, Anton Mashikhin, Anastasia Remizova, Arsenii Ashukha, Aleksei Silvestrov, Naejin Kong, Harshith Goka, Kiwoong Park, and Victor Lempitsky.
\newblock Resolution-robust large mask inpainting with fourier convolutions.
\newblock In \emph{Proceedings of the IEEE/CVF winter conference on applications of computer vision}, pages 2149--2159, 2022.

\bibitem[Tan et~al.(2024)Tan, Liu, Yang, Xue, and Wang]{omnicontrol}
Zhenxiong Tan, Songhua Liu, Xingyi Yang, Qiaochu Xue, and Xinchao Wang.
\newblock Ominicontrol: Minimal and universal control for diffusion transformer.
\newblock \emph{arXiv preprint arXiv:2411.15098}, 2024.

\bibitem[Tan et~al.(2025)Tan, Xue, Yang, Liu, and Wang]{tan2025ominicontrol2}
Zhenxiong Tan, Qiaochu Xue, Xingyi Yang, Songhua Liu, and Xinchao Wang.
\newblock Ominicontrol2: Efficient conditioning for diffusion transformers.
\newblock \emph{arXiv preprint arXiv:2503.08280}, 2025.

\bibitem[Wang et~al.(2018)Wang, Zheng, Liang, Chen, and Yang]{2018Toward}
Bochao Wang, Huabin Zheng, Xiaodan Liang, Yimin Chen, and Meng Yang.
\newblock Toward characteristic-preserving image-based virtual try-on network.
\newblock In \emph{15th European Conference, Munich, Germany, September 8-14, 2018, Proceedings, Part XIII}, 2018.

\bibitem[Wang et~al.(2023)Wang, Bao, Zhou, Wang, Hu, Chen, and Li]{wang2023dire}
Zhendong Wang, Jianmin Bao, Wengang Zhou, Weilun Wang, Hezhen Hu, Hong Chen, and Houqiang Li.
\newblock Dire for diffusion-generated image detection.
\newblock In \emph{Proceedings of the IEEE/CVF International Conference on Computer Vision}, pages 22445--22455, 2023.

\bibitem[Wang et~al.(2004)Wang, Bovik, Sheikh, and Simoncelli]{ssim}
Zhou Wang, Alan~C Bovik, Hamid~R Sheikh, and Eero~P Simoncelli.
\newblock Image quality assessment: from error visibility to structural similarity.
\newblock \emph{IEEE transactions on image processing}, 13\penalty0 (4):\penalty0 600--612, 2004.

\bibitem[Xiao et~al.(2024)Xiao, Wang, Zhou, Yuan, Xing, Yan, Wang, Huang, and Liu]{DBLP:24OmniGen}
Shitao Xiao, Yueze Wang, Junjie Zhou, Huaying Yuan, Xingrun Xing, Ruiran Yan, Shuting Wang, Tiejun Huang, and Zheng Liu.
\newblock Omnigen: Unified image generation.
\newblock \emph{CoRR}, 2024.

\bibitem[Xie et~al.(2023)Xie, Huang, Dong, Zhao, Dong, Zhang, Zhu, and Liang]{DBLP:23GPvton}
Zhenyu Xie, Zaiyu Huang, Xin Dong, Fuwei Zhao, Haoye Dong, Xijin Zhang, Feida Zhu, and Xiaodan Liang.
\newblock {GP-VTON:} towards general purpose virtual try-on via collaborative local-flow global-parsing learning.
\newblock In \emph{{IEEE/CVF} Conference on Computer Vision and Pattern Recognition, {CVPR} 2023, Vancouver, BC, Canada, June 17-24, 2023}, 2023.

\bibitem[Xu et~al.(2025)Xu, Gu, Chen, and Chen]{xu2025ootdiffusion}
Yuhao Xu, Tao Gu, Weifeng Chen, and Arlene Chen.
\newblock Ootdiffusion: Outfitting fusion based latent diffusion for controllable virtual try-on.
\newblock In \emph{Proceedings of the AAAI Conference on Artificial Intelligence}, volume~39, pages 8996--9004, 2025.

\bibitem[Yang et~al.(2023)Yang, Gu, Zhang, Zhang, Chen, Sun, Chen, and Wen]{DBLP:23PaintbyExample}
Binxin Yang, Shuyang Gu, Bo~Zhang, Ting Zhang, Xuejin Chen, Xiaoyan Sun, Dong Chen, and Fang Wen.
\newblock Paint by example: Exemplar-based image editing with diffusion models.
\newblock In \emph{{IEEE/CVF} Conference on Computer Vision and Pattern Recognition, {CVPR} 2023, Vancouver, BC, Canada, June 17-24, 2023}. {IEEE}, 2023.

\bibitem[Yang et~al.(2024)Yang, Jia, Li, and Song]{human_parsing_survey}
Lu~Yang, Wenhe Jia, Shan Li, and Qing Song.
\newblock Deep learning technique for human parsing: A survey and outlook.
\newblock \emph{International Journal of Computer Vision}, 132\penalty0 (8):\penalty0 3270--3301, 2024.

\bibitem[Ye et~al.(2023)Ye, Zhang, Liu, Han, and Yang]{2023IP}
Hu~Ye, Jun Zhang, Sibo Liu, Xiao Han, and Wei Yang.
\newblock Ip-adapter: Text compatible image prompt adapter for text-to-image diffusion models.
\newblock \emph{CoRR}, 2023.

\bibitem[Zhang et~al.(2023)Zhang, Rao, and Agrawala]{DBLP:23ControlNet}
Lvmin Zhang, Anyi Rao, and Maneesh Agrawala.
\newblock Adding conditional control to text-to-image diffusion models.
\newblock In \emph{{IEEE/CVF} International Conference on Computer Vision, {ICCV} 2023, Paris, France, October 1-6, 2023}, 2023.

\bibitem[Zhang et~al.(2018)Zhang, Isola, Efros, Shechtman, and Wang]{lpips}
Richard Zhang, Phillip Isola, Alexei~A Efros, Eli Shechtman, and Oliver Wang.
\newblock The unreasonable effectiveness of deep features as a perceptual metric.
\newblock In \emph{Proceedings of the IEEE conference on computer vision and pattern recognition}, pages 586--595, 2018.

\bibitem[Zhang et~al.(2024)Zhang, Song, Zhan, Chang, Zeng, Chen, Luo, and Liu]{zhang2024boow}
Xuanpu Zhang, Dan Song, Pengxin Zhan, Tianyu Chang, Jianhao Zeng, Qingguo Chen, Weihua Luo, and Anan Liu.
\newblock Boow-vton: Boosting in-the-wild virtual try-on via mask-free pseudo data training.
\newblock \emph{arXiv preprint arXiv:2408.06047}, 2024.

\bibitem[Zhang et~al.(2025)Zhang, Yuan, Song, Wang, and Liu]{easycontrol}
Yuxuan Zhang, Yirui Yuan, Yiren Song, Haofan Wang, and Jiaming Liu.
\newblock Easycontrol: Adding efficient and flexible control for diffusion transformer.
\newblock \emph{arXiv preprint arXiv:2503.07027}, 2025.

\bibitem[Zhao et~al.(2023)Zhao, Chen, Chen, Bao, Hao, Yuan, and Wong]{DBLP:23UniControlNet}
Shihao Zhao, Dongdong Chen, Yen{-}Chun Chen, Jianmin Bao, Shaozhe Hao, Lu~Yuan, and Kwan{-}Yee~K. Wong.
\newblock Uni-controlnet: All-in-one control to text-to-image diffusion models.
\newblock In \emph{Advances in Neural Information Processing Systems 36: Annual Conference on Neural Information Processing Systems 2023, NeurIPS 2023, New Orleans, LA, USA, December 10 - 16, 2023}, 2023.

\bibitem[Zhong et~al.(2023)Zhong, Xu, Qian, and Zhang]{zhong2023rich}
Nan Zhong, Yiran Xu, Zhenxing Qian, and Xinpeng Zhang.
\newblock Rich and poor texture contrast: A simple yet effective approach for ai-generated image detection.
\newblock \emph{CoRR}, 2023.

\bibitem[Zhu et~al.(2023)Zhu, Yang, Zhu, Reda, Chan, Saharia, Norouzi, and Kemelmacher{-}Shlizerman]{DBLP:23TryOnDiffusion}
Luyang Zhu, Dawei Yang, Tyler Zhu, Fitsum Reda, William Chan, Chitwan Saharia, Mohammad Norouzi, and Ira Kemelmacher{-}Shlizerman.
\newblock Tryondiffusion: {A} tale of two unets.
\newblock In \emph{{IEEE/CVF} Conference on Computer Vision and Pattern Recognition, {CVPR} 2023, Vancouver, BC, Canada, June 17-24, 2023}, 2023.

\bibitem[Zhuang et~al.(2024)Zhuang, Zeng, Liu, Yuan, and Chen]{powerpaint}
Junhao Zhuang, Yanhong Zeng, Wenran Liu, Chun Yuan, and Kai Chen.
\newblock A task is worth one word: Learning with task prompts for high-quality versatile image inpainting.
\newblock In \emph{European Conference on Computer Vision}, pages 195--211. Springer, 2024.

\end{thebibliography}
